\documentclass[10pt,twocolumn,letterpaper]{article}

\usepackage[pagenumbers]{cvpr} 

\usepackage[dvipsnames]{xcolor}

\definecolor{cvprblue}{rgb}{0.21,0.49,0.74}
\usepackage[pagebackref,breaklinks,colorlinks,citecolor=cvprblue]{hyperref}

\usepackage{graphicx}
\usepackage{amsmath}
\usepackage{amssymb}
\usepackage{booktabs}
\usepackage[accsupp]{axessibility}
\usepackage[ruled,vlined]{algorithm2e}
\usepackage{float}
\usepackage{xcolor}
\usepackage{colortbl}
\usepackage{amssymb}
\usepackage{makecell}
\usepackage{tablefootnote}
\usepackage{multirow}
\usepackage{pifont}
\usepackage[font=small]{caption}
\usepackage{subcaption}
\usepackage{enumerate}
\usepackage[misc]{ifsym}
\usepackage{arydshln}
\usepackage{bm}

\definecolor{rowblue}{RGB}{198,234,251}   
\definecolor{rowgreen}{RGB}{209,239,191}   
\definecolor{rowgray}{RGB}{245,245,245}   
\definecolor{highgreen}{RGB}{0,139,69}
\definecolor{commentcolor}{RGB}{237,2,140}   

\newcommand{\midsepremove}{\aboverulesep = -0.16mm \belowrulesep = 0mm}
\midsepremove
\newcommand{\midsepdefault}{\aboverulesep = 0.605mm \belowrulesep = 0.984mm}
\midsepdefault

\def\paperID{211} 
\def\confName{CVPR}
\def\confYear{2025}

\title{Seeing What Matters: Empowering CLIP with Patch Generation-to-Selection}

\author{Gensheng Pei$^{1}$, Tao Chen$^{1}$, Yujia Wang$^{2}$, Xinhao Cai$^{1}$, Xiangbo Shu$^{1}$, Tianfei Zhou$^{3}$, Yazhou Yao$^{1}$\thanks{Corresponding author.} \\
\small{$^{1}$Nanjing University of Science and Technology, $^{2}$Zhejiang Sci-Tech University, $^{3}$Beijing Institute of Technology} \\
\small{\url{https://github.com/NUST-Machine-Intelligence-Laboratory/CLIP-PGS}}\\
}

\begin{document}
\maketitle
\begin{abstract}
The CLIP model has demonstrated significant advancements in aligning visual and language modalities through large-scale pre-training on image-text pairs, enabling strong zero-shot classification and retrieval capabilities on various domains. However, CLIP’s training remains computationally intensive, with high demands on both data processing and memory. To address these challenges, recent masking strategies have emerged, focusing on the selective removal of image patches to improve training efficiency. Although effective, these methods often compromise key semantic information, resulting in suboptimal alignment between visual features and text descriptions.
In this work, we present a concise yet effective approach called Patch Generation-to-Selection (CLIP-PGS) to enhance CLIP’s training efficiency while preserving critical semantic content. Our method introduces a gradual masking process in which a small set of candidate patches is first pre-selected as potential mask regions. Then, we apply Sobel edge detection across the entire image to generate an edge mask that prioritizes the retention of the primary object areas. Finally, similarity scores between the candidate mask patches and their neighboring patches are computed, with optimal transport normalization refining the selection process to ensure a balanced similarity matrix.
Our approach, CLIP-PGS, sets new state-of-the-art results in zero-shot classification and retrieval tasks, achieving superior performance in robustness evaluation and language compositionality benchmarks.
\end{abstract}

\begin{figure}
    \begin{center}
        \begin{subfigure}{1.0\linewidth}
            \centering
            \includegraphics[width=\linewidth]{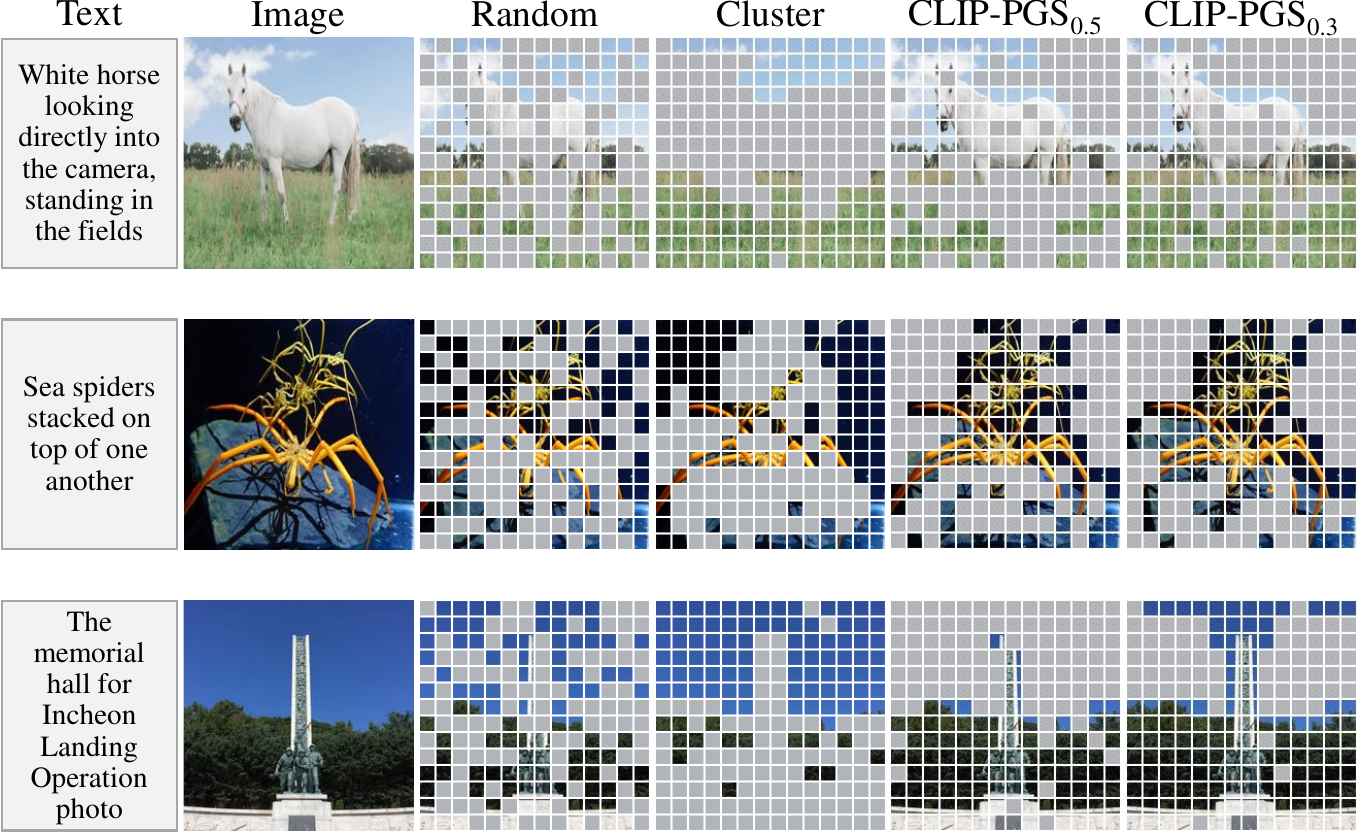}
            \caption{Visualization of Different Masking Strategies}
            \label{fig:1_a}
        \end{subfigure}
        \hfill
        \begin{subfigure}{1.0\linewidth}
            \centering
            \includegraphics[width=\linewidth]{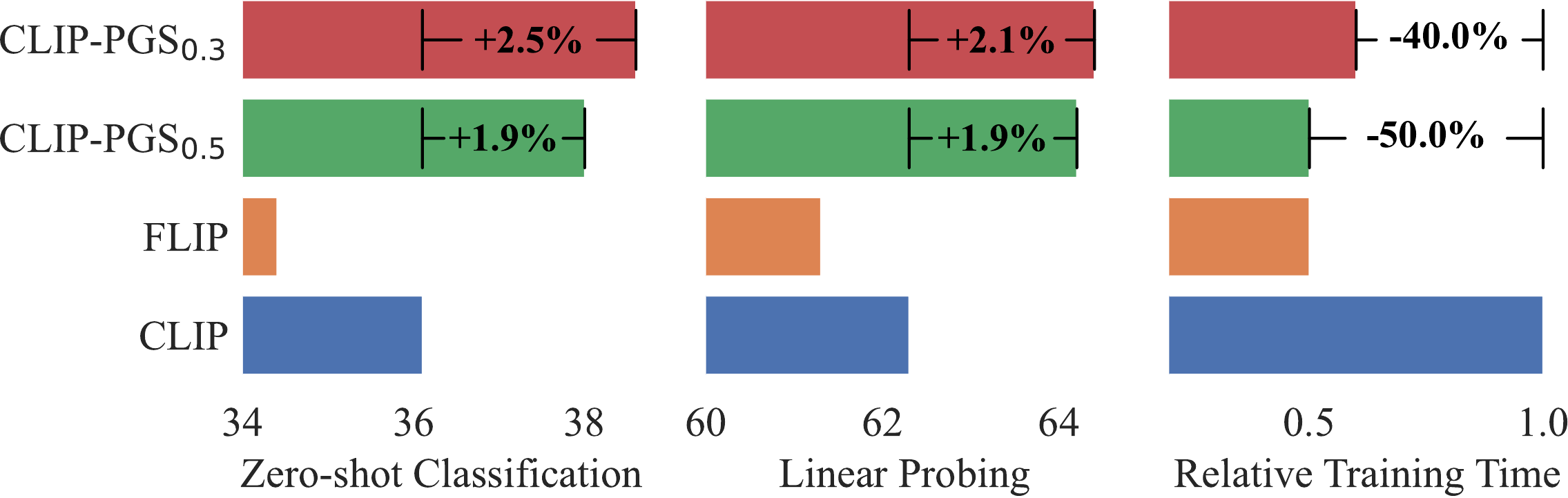}
            \caption{Accuracy (\%) and Relative Training Time ($\times$)}
            \label{fig:1_b}
        \end{subfigure}
    \end{center}
    \vspace{-0.6cm}
    \caption{
    \textbf{Advantages of CLIP-PGS}. (a) Visual comparison of masking strategies: random masking (\eg, FLIP~\cite{flip}), cluster-based masking (\eg, E-CLIP~\cite{e-clip}), and our proposed CLIP-PGS. (b) Improvements in zero-shot classification and linear probing tasks, and relative training time reduction achieved by CLIP-PGS.
    }
    \label{fig:1_motivation}
    \vspace{-0.3cm}
\end{figure}

\begin{figure*}
    \centering
        \includegraphics[width=0.965\linewidth]{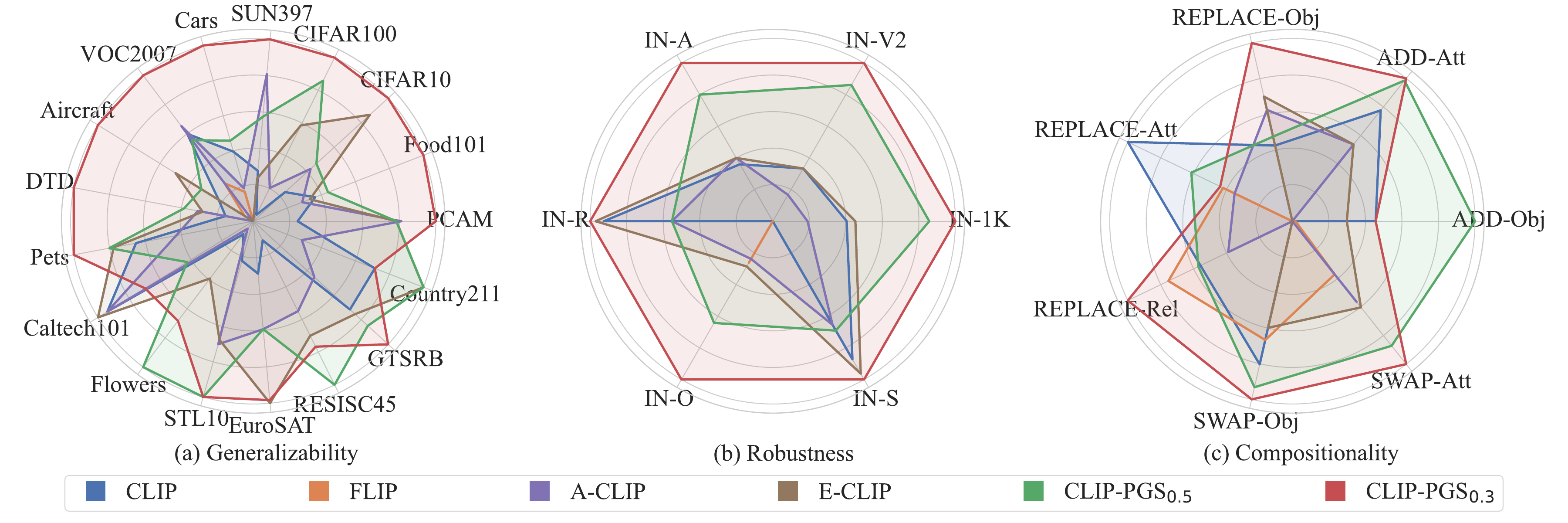}
    \vspace{-0.38cm}
    \caption{\textbf{Performance comparison of vision-language pre-training models}, such as CLIP~\cite{clip}, FLIP~\cite{flip}, A-CLIP~\cite{a-clip}, E-CLIP~\cite{e-clip}, and CLIP-PGS, evaluated across three dimensions using normalized scores: (a) generalizability, (b) robustness, and (c) compositionality.}
    \label{fig:2_radar}
    \vspace{-0.5cm}
\end{figure*}

\vspace{-0.05cm}
\section{Introduction}
The rise of large-scale vision-language models (VLMs)~\cite{gpt2,chatgpt,learning-vlm,vl-bert,li2021align,visualbert,lxmert,mode-clip,synthclip,eva-clip,declip} has revolutionized the field of visual representation learning. Pioneering works such as CLIP~\cite{clip} and ALIGN~\cite{align} have demonstrated the potential of contrastive learning to align visual features with natural language descriptions. By leveraging massive datasets~\cite{cc3m,cc12m,mobileclip,openclip,yfcc100m,datacomp} composed of image-text pairs collected from the internet, these models learn powerful and transferable visual representations that can be applied to a wide range of downstream tasks~\cite{alpha-clip,clipself,tga-zsr,dpe-clip,yao2021jo-src,pei2022hfan,yao2021non-salient,cai2024pkinet} without task-specific fine-tuning.
For example, CLIP uses a simple yet effective dual-encoder architecture that processes images and text independently and aligns them using contrastive loss. This approach has enabled zero-shot classification, showing competitive performance across various datasets without additional data-specific training.

However, models in vision-language pre-training face the ongoing challenge of high computational costs. Their reliance on vast datasets and complex objectives necessitates extensive GPU resources, making efficient training difficult. Recent research has thus aimed to improve efficiency while maintaining competitive performance, introducing innovative strategies in data augmentation~\cite{declip,clip-rocket}, masking~\cite{flip,clipa,a-clip}, and architectural refinement~\cite{mobileclip,detailclip}.

Within computer vision, masked image modeling (MIM) has emerged as a popular self-supervised learning strategy, exemplified by models such as Masked Autoencoders (MAE)~\cite{he2022mae} and BEiT~\cite{beit}.
These models mask parts of the image input, guiding the model to reconstruct the obscured content and, in turn, acquire robust feature representations. 
Inspired by the efficiency gains of MIM, vision-language models have adapted masking techniques to balance computational efficiency and representational quality.
Addressing CLIP's high computational demands, recent developments like FLIP~\cite{flip} introduced random patch masking, allowing the model to process more sample pairs within the same training period by masking 50\%$\sim$75\% of image patches, enhancing scalability for large-scale datasets. Building on these foundations, MaskCLIP~\cite{maskclip} integrates masked image modeling with contrastive language-image training, using self-distillation to align local patch features with semantic text descriptions, which strengthens generalization and transferability.
Further advances include A-CLIP~\cite{a-clip}, which builds on FLIP by introducing an adaptive masking mechanism. A-CLIP retains only the image tokens most semantically related to the paired text, reducing the negative effects of random token removal and ensuring that retained patches maintain meaningful context for visual-text alignment. E-CLIP\cite{e-clip} refines this approach by clustering visually similar patches for masking, preserving or removing coherent visual structures as a group.

Despite the progress made by recent masking strategies in vision-language pre-training, these models still face limitations that may compromise semantic alignment and representation quality. Random masking, as in FLIP~\cite{flip}, can inadvertently remove critical image content, while attention-based methods like A-CLIP~\cite{a-clip} often require additional computational modules, adding to the training complexity. Cluster-based masking in E-CLIP~\cite{e-clip} helps preserve coherent structures but lacks the granularity needed to selectively retain primary semantic regions, and it may unintentionally obscure regions aligned with textual descriptions.

To address these challenges, we propose CLIP-PGS, a concise masking strategy designed to enhance CLIP's training efficiency while carefully preserving essential semantic content, as shown in \cref{fig:1_a}. CLIP-PGS uses a gradual generation-to-selection process: it begins by pre-selecting candidate patches, applies Sobel edge detection to prioritize primary object areas, and then dynamically refines the selection with optimal transport normalization based on similarity scores.
Our method minimizes semantic loss while optimizing training efficiency, resulting in superior performance (see \cref{fig:1_b,fig:2_radar}) across zero-shot classification, retrieval tasks, robustness evaluation, and language compositionality benchmarks. CLIP-PGS achieves efficient training without sacrificing the quality of vision-language alignment, demonstrating that a carefully structured masking approach can enhance both efficiency and semantic integrity.

\section{Related Work}
\noindent\textbf{Vision-Language Pre-training.}
Recently, vision-language models~\cite{chatgpt,vlm-survey,learning-vlm,minigpt,vl-bert,li2021align,visualbert,lxmert} have made significant strides by aligning visual and textual semantics through large-scale pre-training on image-text datasets. Foundational models like CLIP~\cite{clip} and ALIGN~\cite{align} leverage vast internet-sourced image-text pairs, achieving impressive zero-shot classification and retrieval performance across tasks. Despite their success, these models demand substantial computational resources. To enhance efficiency, models like SLIP~\cite{slip}, DeCLIP~\cite{declip}, and MobileCLIP~\cite{mobileclip} integrate self-supervised learning, reducing dependence on supervised data and boosting robustness. FILIP~\cite{filip} enhances local similarity matching for finer cross-modal alignment, while more expressive models such as CoCa~\cite{coca}, BLIP~\cite{blip}, and LaCLIP~\cite{laclip} introduce decoders or captioning modules to enhance language generation, though often with added complexity.

\noindent\textbf{Empowering CLIP with Masking.}
Masked image modeling~\cite{beit,he2022mae,SimMIM,assran2022msn,fang2023eva,liu2023improving,pei2024videomac} is a pivotal technique in computer vision that trains models to reconstruct masked image regions, reducing computational load and enhancing data efficiency by guiding models to concentrate on essential information within constrained inputs. Building on MIM's strengths, recent approaches~\cite{flip,clipa,maskclip,a-clip,e-clip,glip} adapt masking strategies to improve CLIP's efficiency while preserving semantic integrity. FLIP~\cite{flip} introduces random masking of image patches, balancing faster training with accuracy, and establishing an efficient baseline in vision-language pre-training. Recently, more advanced approaches include MaskCLIP~\cite{maskclip}, which employs masked self-distillation to align local patch features with global semantics, strengthening robustness and transferability across tasks. A-CLIP~\cite{a-clip} incorporates attention-based masking to retain tokens aligned with the paired text. Subsequently, E-CLIP~\cite{e-clip} uses cluster-based masking to preserve coherent visual structures, optimizing both efficiency and context preservation.
However, existing methods face two primary challenges: attention-based masking tends to add computational overhead, while cluster-based masking risks inadvertently masking regions corresponding to textual descriptions within the image.
We propose CLIP-PGS, a streamlined approach that enhances pre-training efficiency while selectively preserving critical semantic regions, achieving improved performance with lower computational demands.

\section{Method}
\subsection{Preliminaries}
In the CLIP framework~\cite{clip}, visual and textual representations are aligned directly from image-text pairs using a dual-encoder setup. The image encoder $\mathcal{F}_{v}$ and text encoder $\mathcal{F}_{t}$ independently process an image $\mathcal{I}$ and text $\mathcal{T}$, projecting them into a shared embedding space with L2 normalization. To optimize alignment, CLIP employs the InfoNCE loss~\cite{cl_loss}, which encourages matched image-text pairs to be similar while pushing mismatched pairs apart. For each batch of pairs, the image encoder's loss is defined as the negative log-likelihood of matched pairs normalized by the total similarity with all text embeddings in the batch, scaled by a temperature parameter. A symmetrical loss is applied to the text encoder, and the final loss $\mathcal{L}_{cl}$ averages both, aligning image and text embeddings effectively.

\subsection{CLIP-PGS}

\begin{figure}
    \begin{center}
        \includegraphics[width=1.0\linewidth]{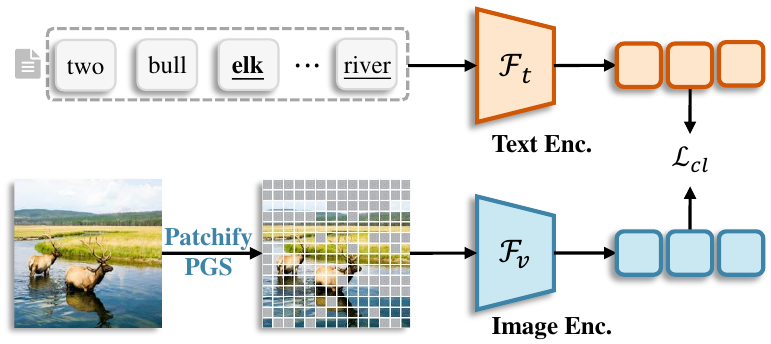}
    \end{center}
    \vspace{-0.6cm}
    \caption{
    \textbf{An illustration of CLIP-PGS}. The text input is processed by the text encoder \(\mathcal{F}_t\), while the image undergoes our patch generation-to-selection strategy before entering the image encoder \(\mathcal{F}_v\). \(\mathcal{L}_{cl}\) subsequently aligns the visual and textual embeddings, strengthening cross-modal representation alignment.
    }
    \label{fig:3_pipeline}
    \vspace{-0.6cm}
\end{figure}

We introduce CLIP-PGS, a simple and efficient masking approach designed to meet the computational demands and semantic preservation challenges inherent in VLMs. CLIP-PGS selectively retains semantically meaningful content, enhancing training efficiency without compromising alignment quality between visual and textual representations.

\noindent\textbf{Gradual Process with Dynamic Masking Ratios.}
Our approach starts by identifying candidate patches for masking. 
We initialize these patches using the same random masking strategy as FLIP\cite{flip}, but with a reduced masking ratio of 5\% compared to FLIP's 50\%.
This work introduces two variants of CLIP-PGS: CLIP-PGS\(_{0.5}\) and CLIP-PGS\(_{0.3}\), where the subscript denotes the lower limit of the masking ratio. Both share an upper limit of 0.5, following FLIP~\cite{flip}. CLIP-PGS\(_{0.5}\) maintains a fixed 0.5 masking ratio as both its lower and upper limits are set to 0.5, while CLIP-PGS\(_{0.3}\) dynamically adjusts within \([0.3, 0.5]\).

The dynamic masking process integrates edge detection (ED) and optimal transport normalization (OTN):
(\textit{i}) We compute cosine similarities between patches to create a similarity matrix that combines feature-based and image-based affinities. 
(\textit{ii}) OTN iteratively refines the similarity matrix to satisfy the doubly stochastic constraint, enhancing patch importance allocation.
(\textit{iii}) ED preserves critical boundaries, minimizing the masking of semantically significant regions.
(\textit{iv}) Patches are ranked by similarity scores to enforce the specified masking bounds, retaining essential patches to meet the lower limit while avoiding excessive masking beyond the upper limit.
Next, we will provide a detailed explanation of the proposed method.

\begin{table*}[t]
\centering
\midsepremove
\setlength{\tabcolsep}{3.0pt}
\resizebox{1.0\linewidth}{!}{
\begin{tabular}{lc||ccccccccccccccccc||c}
\hline\hline
Method & R.T.T. & \rotatebox{65}{Food101} & \rotatebox{65}{CIFAR10} & \rotatebox{65}{CIFAR100} & \rotatebox{65}{SUN397} & \rotatebox{65}{Cars} & \rotatebox{65}{VOC2007} & \rotatebox{65}{Aircraft} & \rotatebox{65}{DTD} & \rotatebox{65}{OxfordPets} & \rotatebox{65}{Caltech101} & \rotatebox{65}{Flowers} & \rotatebox{65}{STL10} & \rotatebox{65}{EuroSAT} & \rotatebox{65}{RESISC45} & \rotatebox{65}{GTSRB} & \rotatebox{65}{Country211} & \rotatebox{65}{PCam} & \rotatebox{65}{Average} \\
\midrule
CLIP~\cite{clip} & 1.0$\times$   & 42.3 & 57.7 & 25.0 & 44.1 & 17.0 & 50.5 & 1.7 & 16.5 & 53.9 & \underline{73.5} & 26.0 & 82.0 & 18.7 & 26.5 & 9.4 & \underline{4.5} & 48.0 & 35.1 \\ 
FLIP~\cite{flip} & \textbf{0.5}$\times$   & 39.9 & 52.8 & 24.5 & 42.8 & 15.9 & 46.6 & 1.4 & 15.9 & 46.0 & 70.4 & 25.3 & 80.2 & 17.0 & 25.8 & 5.6 & 4.0 & 47.1 & 33.0 \\ 
A-CLIP~\cite{a-clip} & 1.1$\times$ & 41.8 & 61.6 & 27.1 & \underline{46.6} & 16.0 & \underline{51.1} & 1.3 & 17.1 & 51.2 & \underline{73.5} & 25.7 & \underline{85.8} & 20.5 & 29.1 & 8.0 & 4.2 & \underline{50.1} & 35.9 \\ 
E-CLIP~\cite{e-clip} & \underline{0.6}$\times$ & 42.1 & \underline{70.7} & 32.0 & 43.9 & 15.1 & 43.6 & \underline{2.2} & 17.0 & 55.4 & \textbf{73.7} & 28.4 & 85.6 & \textbf{22.9} & 30.0 & 9.6 & \textbf{4.7} & 50.0 & 36.9 \\ 
\midrule
\rowcolor{rowgray} \textit{Ours} &&&&&&&&&&&&&&&&&&&  \\ 
\rowcolor{rowblue} \textbf{CLIP-PGS}$_{0.5}$ & \textbf{0.5}$\times$ & \underline{42.8} & 62.5 & \underline{35.5} & 45.5 & \underline{17.3} & 50.0 & 1.9 & \underline{17.4} & \underline{55.7} & 71.8 & \textbf{33.2} & \textbf{88.2} & 20.5 & \textbf{31.8} & \underline{10.1} & \textbf{4.7} & 50.0 & \underline{37.6} \\ 
\rowcolor{rowblue} \textbf{CLIP-PGS}$_{0.3}$ & \underline{0.6}$\times$ & \textbf{46.5} & \textbf{73.5} & \textbf{37.3} & \textbf{47.5} & \textbf{19.9} & \textbf{55.1} & \textbf{3.1} & \textbf{19.8} & \textbf{58.1} & 72.7 & \underline{30.7} & \textbf{88.2} & \underline{22.8} & \underline{30.4} & \textbf{10.9} & \underline{4.5} & \textbf{50.8} & \textbf{39.5} \\ 
\hline\hline
\end{tabular}}
\midsepdefault
\vspace{-0.2cm}
\caption{\textbf{Zero-shot classification results}. We evaluate performance on 17 diverse classification datasets, reporting both top-1 accuracy (\%) and the overall average. The optimal result is highlighted in \textbf{bold}, and the second-best result is \underline{underlined}. The training time, represented as Relative Training Time (R.T.T.), is benchmarked against CLIP's and set as 1.0$\times$, with other methods shown in relation to this.}
\label{tab:1_zsc}
\vspace{-0.2cm}
\end{table*}

\noindent\textbf{Edge Detection.}
In CLIP-PGS, edge detection (Sobel~\cite{sobel_ed} as used in this work, see \S\ref{sec:ablation} for ablation details) is applied to the whole image to create an edge map that emphasizes prominent object boundaries and contours. This edge map plays a crucial role in preserving critical semantic information during the masking process. When identifying candidate patches for masking, the edge map is utilized to assign higher importance to patches near strong edges, thereby reducing the likelihood of obscuring key regions such as object outlines and high-contrast details. 

Specifically, if a patch is initially marked for removal but exhibits high edge scores, it is retained, while patches with weak edge signals and low candidate heuristics are more likely to be masked. This strategic integration of global edge detection with candidate selection ensures that semantically meaningful areas are preserved, maintaining the alignment between visual features and their textual descriptions while adhering to the desired masking ratio.

\noindent\textbf{Optimal Transport Normalization.}
To refine the selection of masked regions and preserve critical semantic information, CLIP-PGS leverages optimal transport normalization (OTN, see \cref{tab:6_ablation}), implemented via the Sinkhorn algorithm~\cite{sinkhorn_ot}, to process similarity scores between patches. 

The process begins with computing cosine similarity between patches to form a similarity matrix $\bm{S}$, where each entry $\bm{S}_{ij}$ denotes the similarity between patch $i$ and patch $j$. This similarity is calculated using $\bm{S} = \bm{X} \bm{X}^\top$, with $\bm{X}$ being the normalized embedding matrix ($\bm{X} = \bm{X} / \|\bm{X}\|_2$).  
The final similarity matrix is computed by integrating both feature and image similarities through a weighted sum: $\bm{S} = \alpha \bm{S}_x + (1 - \alpha) \bm{S}_I$, where $\bm{S}_x$ and $\bm{S}_I$ are the cosine similarities of features and images, respectively, and $\alpha$ is adjusted based on the training epoch. 
The Sinkhorn algorithm is then applied to $\bm{S}$ to iteratively normalize its rows and columns, resulting in a doubly stochastic matrix that ensures a balanced distribution of similarity scores. 

The updated similarity matrix $\bm{S}^\prime = \bm{S} + \texttt{Sinkhorn}(\bm{S})$ is crucial for guiding the masking process.
OTN uses this balanced similarity matrix to distribute attention across patches, retaining those with high similarity to adjacent regions, thereby preventing the loss of essential visual cues. This balanced selection, facilitated by OTN, not only preserves critical features but also maintains robust alignment between visual and textual modalities during training.

\noindent\textbf{Performance Benefits.}
CLIP-PGS boosts pre-training efficiency while preserving semantic integrity, yielding high performance in zero-shot classification, retrieval, robustness, and language composition tasks. See \S\ref{sec:exp} for details.

\begin{table*}[t]
\centering
\midsepremove
\setlength{\tabcolsep}{3.2pt}
\resizebox{1.0\linewidth}{!}{
\begin{tabular}{l||ccccccccc||ccccccccc}
\hline\hline
& \multicolumn{9}{c||}{Text Retrieval} & \multicolumn{9}{c}{Image Retrieval} \\
\cmidrule(lr){2-10} \cmidrule(lr){11-19}
Method & \multicolumn{3}{c}{MS-COCO} & \multicolumn{3}{c}{Flickr8K} & \multicolumn{3}{c||}{Flickr30K} & \multicolumn{3}{c}{MS-COCO} & \multicolumn{3}{c}{Flickr8K} & \multicolumn{3}{c}{Flickr30K}\\
\cmidrule(lr){2-4} \cmidrule(lr){5-7} \cmidrule(lr){8-10} \cmidrule(lr){11-13} \cmidrule(lr){14-16} \cmidrule(lr){17-19}
& R@1 & R@5 & R@10 & R@1 & R@5 & R@10 & R@1 & R@5 & R@10 & R@1 & R@5 & R@10 & R@1 & R@5 & R@10 & R@1 & R@5 & R@10 \\
\midrule
CLIP~\cite{clip} & 34.6 & \underline{62.0} & 72.7 & 55.7 & 81.6 & 89.9 & \underline{58.5} & \underline{83.8} & 89.1 & 23.5 & 47.8 & 59.7 & 40.5 & 68.9 & 80.2 & 43.2 & 70.4 & 80.4 \\
FLIP~\cite{flip} & 32.6 & 59.1 & 70.6 & 55.0 & 80.9 & 88.9 & 53.8 & 80.8 & 88.5 & 22.6 & 46.1 & 58.1 & 40.3 & 68.1 & 78.6 & 41.5 & 67.9 & 77.5 \\
A-CLIP~\cite{a-clip} & 33.7 & 60.2 & 71.0 & 53.7 & 80.1 & 88.0 & 55.3 & 81.4 & 87.6 & 23.9 & 48.3 & 60.0 & 40.6 & 68.9 & 78.9 & 43.1 & 70.1 & 78.8 \\
E-CLIP~\cite{e-clip} & 34.3 & \underline{62.0} & \underline{73.3} & 57.0 & 82.7 & 90.1 & 55.8 & \textbf{84.2} & 89.6 & 23.8 & 48.2 & 59.8 & 42.0 & 69.4 & 79.6 & 43.3 & 70.9 & 80.2 \\
\midrule
\rowcolor{rowgray} \textit{Ours} &&&&&&&&&&&&&&&&&&  \\ 
\rowcolor{rowblue} \textbf{CLIP-PGS}$_{0.5}$ & \underline{35.2} & 61.9 & 72.8 & \textbf{58.5} & \textbf{83.6} & \underline{90.6} & 57.7 & 82.7 & \underline{90.4} & \underline{24.3} & \underline{48.8} & \underline{60.5} & \underline{43.5} & \underline{70.7} & \underline{81.0} & \underline{45.3} & \underline{72.9} & \underline{81.2} \\
\rowcolor{rowblue} \textbf{CLIP-PGS}$_{0.3}$ & \textbf{36.0} & \textbf{64.4} & \textbf{74.6} & \underline{58.3} & \underline{82.9} & \textbf{90.8} & \textbf{59.9} & 83.5 & \textbf{90.8} & \textbf{25.1} & \textbf{49.5} & \textbf{61.6} & \textbf{44.4} & \textbf{71.7} & \textbf{81.1} & \textbf{47.1} & \textbf{73.5} & \textbf{82.0} \\
\hline\hline
\end{tabular}}
\midsepdefault
\vspace{-0.2cm}
\caption{\textbf{Zero-shot text/image retrieval results}. We evaluate performance on the MS-COCO~\cite{ms-coco}, Flickr8k~\cite{flickr}, and Flickr30k~\cite{flickr} datasets, reporting Recall@1 (\%, R@1), Recall@5 (\%, R@5), and Recall@10 (\%, R@10) for both text and image retrieval tasks.}
\label{tab:2_zsr}
\vspace{-0.2cm}
\end{table*}

\section{Experiments}\label{sec:exp}
In this section, we outline the implementation details (\S\ref{sec:imp_setup}) and present comparison results (\S\ref{sec:exp_results}) to verify the proposed method's generalizability on multiple datasets and its robustness to out-of-distribution scenarios. Additionally, we report the training efficiency of existing methods and conduct ablation studies (\S\ref{sec:ablation}) to analyze the proposed components and design choices systematically.

\subsection{Implementation}\label{sec:imp_setup}
\noindent\textbf{Datasets.}
Our model is trained on the Conceptual Captions 12M (CC12M)~\cite{cc12m} dataset, which comprises approximately 12 million image-text pairs, specifically designed for language-image pre-training. We evaluate all models on 17 diverse classification datasets (\eg, Food101~\cite{food101}, CIFAR~\cite{cifar} and SUN397~\cite{sun397}), retrieval datasets (\eg, MS COCO~\cite{ms-coco} and Flickr~\cite{flickr}), the ImageNet-1K~\cite{deng2009imagenet} dataset and its out-of-distribution variants, as well as a language compositionality dataset (\ie, SugarCrepe~\cite{sugarcrepe}). For additional details on datasets, please refer to the appendix.

\noindent\textbf{Architecture.}
Following the CLIP~\cite{clip} and OpenCLIP~\cite{openclip} frameworks, we utilize the ViT-B/16~\cite{vit} architecture as the backbone for the image encoder and a 12-layer transformer with 512-dimensional embeddings and 8 attention heads as the text encoder. See the appendix for architecture details.

\noindent\textbf{Pre-training.}
In our language-image pre-training model, we resize the image to a resolution of $224 \times 224$, and tokenize the text into 77 tokens using a 49K token vocabulary, with truncation or padding applied as necessary. The class token is then embedded into a 512 dimensional feature vector through a multi-layer perceptron (MLP). We optimize the model using the AdamW optimizer~\cite{adamw}, setting the learning rate to 1e-3, $\beta_{1}$ to 0.9, $\beta_{2}$ to 0.98, and applying a weight decay of 0.2, along with a cosine decay schedule for the learning rate. Our model is trained for 32 epochs with a batch size of 4,096 on 8 NVIDIA TESLA V100 GPUs, using PyTorch’s automatic mixed precision library~\cite{pytorch}.

\noindent\textbf{Downstream Evaluation Tasks.}
We evaluate our model covering \textbf{five} standard benchmark scenarios: zero-shot classification, zero-shot text/image retrieval, linear probing, robustness assessment, and language compositionality, following established evaluation protocols~\cite{clip,slip,openclip,flip}.\footnote{\url{https://github.com/LAION-AI/CLIP_benchmark}\label{clip-benchmark}}

For the five downstream evaluation tasks outlined above, we adhere strictly to the implementation protocols established in the CLIP baseline~\cite{clip}, ensuring consistency and reliability throughout the evaluation process. This alignment allows for fair comparisons and validates the generalizability and robustness of our methodology.

\noindent\textbf{Baselines.}
We introduce the four baseline models used for comparison, \ie, CLIP~\cite{clip}, FLIP~\cite{flip}, A-CLIP~\cite{a-clip}, and E-CLIP~\cite{e-clip}. We comprehensively compare different masking strategies in language-image pre-training, reporting baseline results directly from original papers for fairness. All experiments are conducted under consistent settings to ensure reliable conclusions. Performance in various downstream tasks is detailed in the following sections.

\subsection{Comparison with SOTA Methods}\label{sec:exp_results}
\noindent\textbf{Zero-Shot Classification.}
\cref{tab:1_zsc} presents the zero-shot classification results for our proposed method, CLIP-PGS, with two variants (CLIP-PGS$_{0.5}$ and CLIP-PGS$_{0.3}$) using lower limit masking rates of 0.5 and 0.3. We evaluate CLIP-PGS on 17 diverse classification datasets, comparing it to state-of-the-art models such as CLIP~\cite{clip}, FLIP~\cite{flip}, A-CLIP~\cite{a-clip}, and E-CLIP~\cite{e-clip}.
Due to dataset distribution differences, model performance varies. Notably, compact feature spaces (\eg, CIFAR-10) are more sensitive to masking, while greater class diversity (\eg, CIFAR-100) mitigates this effect, as seen in FLIP~\cite{flip} and E-CLIP~\cite{e-clip}.

Our CLIP-PGS$_{0.3}$ demonstrates significant improvements on more complex datasets, achieving average accuracy gains of \textbf{6.5\%} and \textbf{3.6\%} over FLIP~\cite{flip} and A-CLIP~\cite{a-clip}, respectively. In particular, our CLIP-PGS$_{0.5}$, with a 0.5 lower limit masking rate, improves accuracy by 4.6\% over FLIP~\cite{flip} while consuming the same training time. Both variants of CLIP-PGS, with training times reduced to 0.5$\times$ and 0.6$\times$ of CLIP's duration, either match or exceed baseline accuracy, underscoring the efficiency of our approach. Additionally, CLIP-PGS$_{0.3}$ outperforms E-CLIP~\cite{e-clip}, showing an average gain of \textbf{2.9\%}, and achieves the highest top-1 accuracy on \textbf{12} out of 17 benchmarking datasets. It performs especially well on challenging benchmarks like Food101~\cite{food101}, SUN397~\cite{sun397}, and VOC2007~\cite{voc2007}, establishing new state-of-the-art results.

These findings confirm CLIP-PGS's generalizability, efficiency, and adaptability, making it well-suited for real-world applications in various task scenarios.

\begin{table}[t]
\vspace{0.2cm}
\centering
\midsepremove
\setlength{\tabcolsep}{8.5pt}
\resizebox{1.0\linewidth}{!}{
\begin{tabular}{l||ccc}
\hline\hline
Method & \rotatebox{0}{CIFAR10} & \rotatebox{0}{CIFAR100} & \rotatebox{0}{ImageNet-1K} \\ 
\midrule
CLIP~\cite{clip}     & 88.0 & 67.4 & 62.3  \\ 
FLIP~\cite{flip}     & 85.9 & 65.5 & 61.3  \\ 
A-CLIP~\cite{a-clip} & 86.4 & 66.1 & 62.0 \\ 
E-CLIP~\cite{e-clip} & 89.0 & 69.7 & 62.7 \\ 
\midrule
\rowcolor{rowgray} \textit{Ours} &&& \\
\rowcolor{rowblue} \textbf{CLIP-PGS}$_{0.5}$ & \underline{89.5} \textcolor{highgreen}{\textbf{(+0.5)}} & \underline{70.3} \textcolor{highgreen}{\textbf{(+0.6)}} & \underline{64.2} \textcolor{highgreen}{\textbf{(+1.5)}} \\
\rowcolor{rowblue} \textbf{CLIP-PGS}$_{0.3}$ & \textbf{90.0} \textcolor{highgreen}{\textbf{(+1.0)}} & \textbf{72.3} \textcolor{highgreen}{\textbf{(+2.6)}} & \textbf{64.4} \textcolor{highgreen}{\textbf{(+1.7)}} \\
\hline\hline
\end{tabular}}
\midsepdefault
\vspace{-0.2cm}
\caption{\textbf{Linear probing classification results}. We evaluate all models on three common datasets, \ie, CIFAR10~\cite{cifar}, CIFAR100~\cite{cifar}, and ImageNet-1K~\cite{deng2009imagenet}, training each for 10 epochs under a consistent linear training setup. We present top-1 accuracy (\%), with gains over the stronger baseline highlighted in \textcolor{highgreen}{\textbf{(green)}}.}
\label{tab:3_lp}
\vspace{-0.2cm}
\end{table}

\noindent\textbf{Zero-Shot Text/Image Retrieval.}
We evaluate the zero-shot text and image retrieval performance of our proposed method, CLIP-PGS, on MS-COCO~\cite{ms-coco}, Flickr8k~\cite{flickr}, and Flickr30k~\cite{flickr}, comparing it to state-of-the-art models. 
\cref{tab:2_zsr} reports Recall@1, Recall@5, and Recall@10 for both text and image retrieval tasks on these datasets.

The CLIP-PGS$_{0.3}$ variant achieves top-tier performance in most metrics, reaching a Recall@1 of \textbf{36.0\%} on MS-COCO text retrieval and \textbf{25.1\%} on MS-COCO image retrieval, outperforming other models in these categories. CLIP-PGS$_{0.5}$ also delivers strong results, securing either the best or second-best scores in multiple retrieval tasks.

In comparison to E-CLIP~\cite{e-clip}, CLIP-PGS$_{0.3}$ consistently achieves higher retrieval accuracy, particularly on challenging tasks, while CLIP-PGS$_{0.5}$ closely follows and even surpasses E-CLIP~\cite{e-clip} on specific metrics. Both variants of CLIP-PGS also show improvements over FLIP~\cite{flip} and A-CLIP~\cite{a-clip}, with substantial gains on Flickr8K and Flickr30K. Our method sets new benchmarks over diverse datasets in zero-shot text and image retrieval.

\noindent\textbf{Linear Probing.}
As shown in \cref{tab:3_lp}, we assess the linear probing performance on three widely used datasets: CIFAR10~\cite{cifar}, CIFAR100~\cite{cifar}, and ImageNet-1K~\cite{deng2009imagenet}.

\begin{table*}[ht]
\centering
\midsepremove
\setlength{\tabcolsep}{3.2pt}
\resizebox{1.0\linewidth}{!}{
\begin{tabular}{l||cccccc||ccc}
\hline\hline
Method & \rotatebox{0}{ImageNet-1K} & \rotatebox{0}{ImageNet-V2} & \rotatebox{0}{ImageNet-A} & \rotatebox{0}{ImageNet-R} & \rotatebox{0}{ImageNet-O} & \rotatebox{0}{ImageNet-Sketch} & \rotatebox{0}{Average} & \rotatebox{0}{ID Average} & \rotatebox{0}{OOD Average} \\ 
\midrule
CLIP~\cite{clip}     & 36.1 & 30.7 & 8.0 & 47.6 & 38.4 & 24.9 & 31.0 & 36.1 & 29.0 \\ 
FLIP~\cite{flip}     & 34.4 & 29.5 & 7.1 & 41.4 & 39.5 & 20.1 & 28.7 & 34.4 & 27.5 \\ 
A-CLIP~\cite{a-clip} & 35.2 & 30.1 & 8.1 & 45.1 & 39.4 & 23.7 & 30.3 & 35.2 & 30.3 \\ 
E-CLIP~\cite{e-clip} & 36.3 & 30.7 & 8.1 & \underline{47.9} & 39.6 & \underline{25.4} & 31.3 & 36.3 & 30.3 \\ 
\midrule
\rowcolor{rowgray} \textit{Ours} &&&&&&&&&  \\
\rowcolor{rowblue} \textbf{CLIP-PGS}$_{0.5}$ & \underline{38.0} & \underline{32.6} & \underline{9.1} & 45.1 & \underline{41.1} & 23.9 & \underline{31.6} & \underline{38.0} & \underline{30.4} \\  
\rowcolor{rowblue} \textbf{CLIP-PGS}$_{0.3}$ & \textbf{38.6} & \textbf{33.1} & \textbf{9.6} & \textbf{48.1} & \textbf{42.6} & \textbf{25.6} & \textbf{32.9} & \textbf{38.6} & \textbf{31.8} \\
\hline\hline
\end{tabular}}
\midsepdefault
\vspace{-0.2cm}
\caption{\textbf{Robustness assessment results}. We evaluate model robustness on ImageNet-1K~\cite{deng2009imagenet} and five of its variants~\cite{imagenet-v2,imagenet-ao,imagenet-r,imagenet-sketch}, reporting top-1 accuracy (\%) along with overall averages for in-distribution (ID) and out-of-distribution (OOD) performance.}
\label{tab:4_robust}
\end{table*}

\begin{table*}[ht]
\centering
\midsepremove
\setlength{\tabcolsep}{7.9pt}
\resizebox{1.0\linewidth}{!}{
\begin{tabular}{lc||ccccccc||ccc}
\hline\hline
\multirow{2}{*}{\begin{tabular}[c]{@{}c@{}}Method\end{tabular}} & \multirow{2}{*}{\begin{tabular}[c]{@{}c@{}}R.T.T.\end{tabular}} &  \multicolumn{3}{c}{REPLACE} & \multicolumn{2}{c}{SWAP} & \multicolumn{2}{c||}{ADD} & \multicolumn{3}{c}{Average} \\
\cmidrule(lr){3-5} \cmidrule(lr){6-7} \cmidrule(lr){8-9} \cmidrule(lr){10-12}
& & Object & Attribute & Relation & Object & Attribute & Object & Attribute & Object & Attribute & Relation \\
\midrule
CLIP~\cite{clip} & 1.0$\times$ & 85.8 & \textbf{79.2} & 64.5 & 61.8 & 58.7 & \underline{74.2} & 68.4 & \underline{73.7} & \underline{68.8} & 64.5 \\
FLIP~\cite{flip} & \textbf{0.5}$\times$ & 84.1 & 75.9 & \underline{66.0} & 60.2 & 61.6 & 71.7 & 63.2 & 72.0 & 66.9 & \underline{66.0} \\
A-CLIP~\cite{a-clip} & 1.1$\times$ & \underline{86.6} & 75.5 & 63.2 & 52.4 & 63.1 & 71.6 & 66.8 & 71.6 & 68.4 & 63.2 \\
E-CLIP~\cite{e-clip} & \underline{0.6}$\times$ & 86.9 & 73.5 & 60.2 & 59.4 & 63.4 & 73.3 & 66.8 & 73.2 & 68.4 & 60.2 \\
\midrule
\rowcolor{rowgray} \textit{Ours} &&&&&&&&&&& \\
\rowcolor{rowblue} \textbf{CLIP-PGS}$_{0.5}$ & \textbf{0.5}$\times$ & 86.0 & \underline{77.0} & 64.6 & 63.3 & \underline{65.5} & \textbf{77.3} & \underline{69.8} & \textbf{75.5} & \textbf{70.8} & 64.6 \\
\rowcolor{rowblue} \textbf{CLIP-PGS}$_{0.3}$ & \underline{0.6}$\times$ & \textbf{88.1} & 76.0 & \textbf{67.9} & \underline{64.1} & \textbf{66.5} & \underline{74.2} & \textbf{69.9} & \textbf{75.5} & \textbf{70.8} & \textbf{67.9} \\
\hline\hline
\end{tabular}}
\midsepdefault
\vspace{-0.2cm}
\caption{\textbf{Language compositionality results}. We evaluate the compositionality of vision-language models on the SugarCrepe~\cite{sugarcrepe} dataset, which tests models by generating mismatched captions by replacing, swapping, or adding fine-grained atomic concepts (object, attribute, and relation). We report Recall@1 (\%) and the overall average for each atomic concept.}
\label{tab:5_composition}
\vspace{-0.2cm}
\end{table*}

Our approach outperforms the baselines, with CLIP-PGS$_{0.3}$ achieving the highest accuracy on all datasets. Notably, CLIP-PGS$_{0.3}$ improves over E-CLIP~\cite{e-clip} by \textbf{1.0\%} on CIFAR10, \textbf{2.6\%} on CIFAR100, and \textbf{1.7\%} on ImageNet-1K. CLIP-PGS$_{0.5}$ also achieves competitive results, surpassing E-CLIP~\cite{e-clip} by \textbf{0.5\%} on CIFAR10, \textbf{0.6\%} on CIFAR100, and \textbf{1.5\%} on ImageNet-1K. The effectiveness of CLIP-PGS in linear probing highlights its superior capability to capture meaningful representations.

\noindent\textbf{Robustness Assessment.}
To evaluate robustness, we test CLIP-PGS on ImageNet-1K~\cite{deng2009imagenet} alongside five of its variants, \ie, ImageNet-V2~\cite{imagenet-v2}, ImageNet-A~\cite{imagenet-ao}, ImageNet-R~\cite{imagenet-r}, ImageNet-O~\cite{imagenet-ao}, and ImageNet-Sketch~\cite{imagenet-sketch}.
We provide the zero-shot top-1 accuracy for each dataset, along with overall in-distribution (ID) and out-of-distribution (OOD) averages, as detailed in \cref{tab:4_robust}.

CLIP-PGS$_{0.3}$ achieves the highest robustness scores over all datasets, reaching top-1 accuracy of \textbf{38.6\%} on ImageNet-1K, \textbf{9.6\%} on ImageNet-A, and \textbf{42.6\%} on ImageNet-O. Its ID and OOD averages are \textbf{32.9\%} and \textbf{31.8\%}, respectively, surpassing all other models. CLIP-PGS$_{0.5}$ shows competitive results, securing the second-best ID and OOD averages at \textbf{31.6\%} and \textbf{30.4\%}, respectively.

Compared to E-CLIP~\cite{e-clip}, which achieves 47.9\% on ImageNet-R and 25.4\% on ImageNet-Sketch, CLIP-PGS$_{0.3}$ provides improved consistency covering all datasets. This consistency reflects the robustness and adaptability of CLIP-PGS in handling diverse distributional shifts.

\noindent\textbf{Language Compositionality.}
We evaluate the language compositionality of CLIP-PGS on the SugarCrepe~\cite{sugarcrepe} dataset. \cref{tab:5_composition} reports Recall@1 for each atomic concept (object, attribute, and relation) in three tasks: replacing, swapping, and adding concepts in captions. Additionally, we provide an overall average for each concept type.

Our method exhibits notable improvements in language compositionality. CLIP-PGS$_{0.3}$ achieving leading scores on most categories. Specifically, CLIP-PGS$_{0.3}$ surpasses A-CLIP~\cite{a-clip} and FLIP~\cite{flip}, with \textbf{88.1\%} in object replacement and \textbf{67.9\%} in relation handling, marking an average gain over competing models. In addition to outperforming E-CLIP~\cite{e-clip}, CLIP-PGS$_{0.3}$ achieves higher accuracy in attribute handling compared to other baselines, indicating its robust performance in diverse concept manipulations.

CLIP-PGS$_{0.5}$ likewise shows notable results, leading in object addition with \textbf{77.3\%} and securing the highest overall averages for object and attribute tasks. Compared to FLIP~\cite{flip}, CLIP-PGS$_{0.5}$ offers an improvement in handling fine-grained attributes and relations. Evaluation results emphasize the effectiveness of CLIP-PGS, especially in tasks requiring nuanced compositionality in vision-language scenarios. Its capabilities make it well-suited for managing complex concept manipulations in real-world applications.

\begin{figure*}[t]
    \centering
        \includegraphics[width=1.0\linewidth]{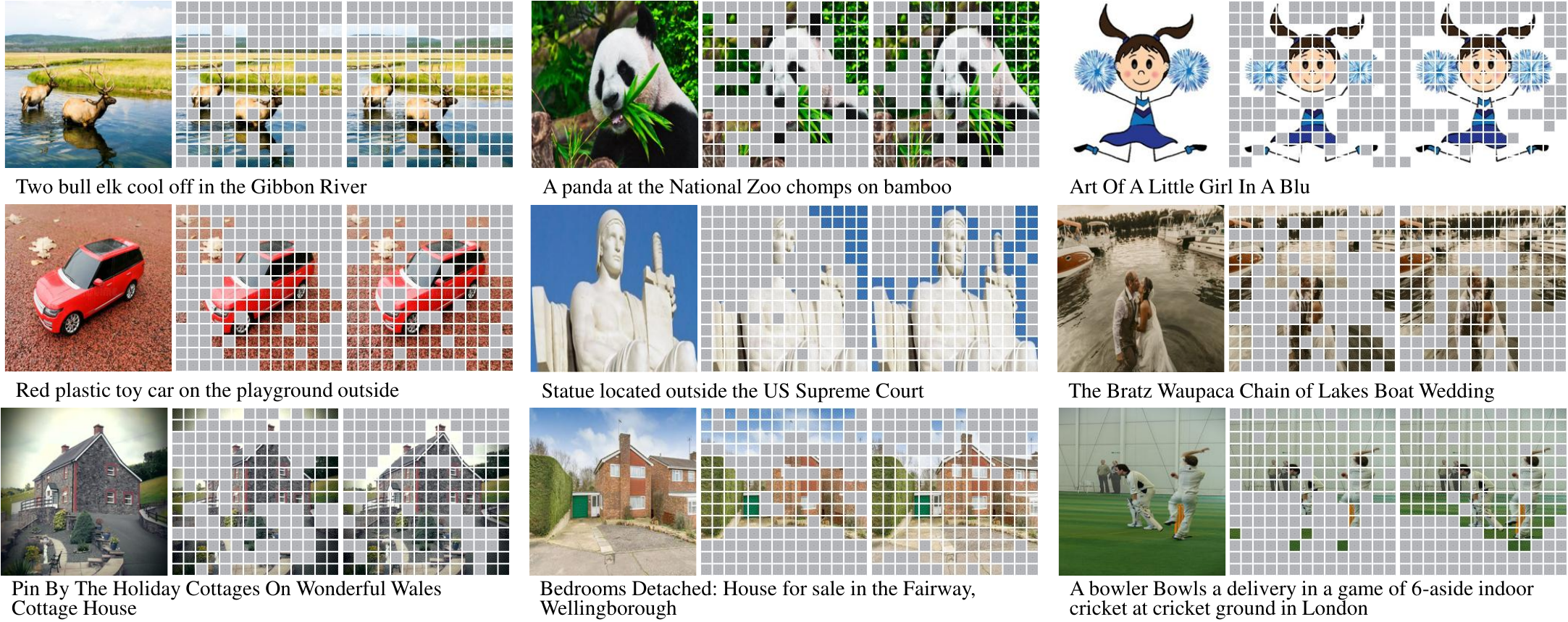}
    \vspace{-0.7cm}
    \caption{\textbf{Visualization of masking regions}. We use ViT-B/16~\cite{vit} as the image encoder, displaying each sample with the text description, the original image (\textbf{left}), and masking results from CLIP-PGS$_{0.5}$ (\textbf{middle}) at a fixed 0.5 masking ratio, and CLIP-PGS$_{0.3}$ (\textbf{right}) with a variable masking ratio between 0.3 and 0.5. Our models effectively retain the visual content relevant to the accompanying text context.}
    \label{fig:4_patch_vis}
    \vspace{-0.1cm}
\end{figure*}

\begin{figure*}[t]
    \centering
    \begin{minipage}{0.71\linewidth}
        \centering
        \midsepremove
        \setlength{\tabcolsep}{5.2pt}
        \resizebox{\linewidth}{!}{
        \begin{tabular}{l||ccccccc}
        \hline\hline
        \multirow{2}{*}{Method} & \multicolumn{3}{c}{Component (extra cost)} & \multicolumn{2}{c}{ImageNet-1K} & \multicolumn{2}{c}{MS-COCO} \\
        \cmidrule(lr){2-4} \cmidrule(lr){5-6} \cmidrule(lr){7-8}
        & \colorbox{red!25}{MR ($<$1.0\%)} & \colorbox{green!25}{ED ($\sim$1\%)} & \colorbox{cyan!25}{OTN ($\sim$1\%)} & ZS & LP & TR & IR \\
        \midrule
        \rowcolor{rowgray} \textcolor{gray}{\textit{Baseline}~\cite{clip}}
                            & \textcolor{gray}{-}     & \textcolor{gray}{-} & \textcolor{gray}{-} & \textcolor{gray}{36.1} & \textcolor{gray}{62.3} & \textcolor{gray}{34.6} & \textcolor{gray}{23.5} \\
        \midrule
        \rowcolor{rowgray} \textcolor{gray}{\textit{Random Mask}~\cite{flip}}
                            & \textcolor{gray}{0.5}   & \textcolor{gray}{-} & \textcolor{gray}{-} & \textcolor{gray}{34.4} & \textcolor{gray}{61.3} & \textcolor{gray}{32.6} & \textcolor{gray}{22.6} \\
        \midrule
        \multirow{4}{*}{\textbf{CLIP-PGS}$_{0.5}$}
                            & 0.5        & \textcolor{red!80}{\ding{55}} & \textcolor{red!80}{\ding{55}} & 35.2 & 61.9 & 33.7 & 22.8 \\
                            & 0.5        & \textcolor{highgreen}{\ding{51}} & \textcolor{red!80}{\ding{55}} & 36.2 & 62.8 & 34.1 & 23.4 \\
                            & 0.5        & \textcolor{red!80}{\ding{55}} & \textcolor{highgreen}{\ding{51}} & 36.3 & 62.7 & 33.9 & 23.2\\
                            & 0.5        & \textcolor{highgreen}{\ding{51}} & \textcolor{highgreen}{\ding{51}} & 38.0 & 64.2 & 35.2 & 24.3 \\
        \midrule
        \multirow{4}{*}{\textbf{CLIP-PGS}$_{0.3}$}
                            & [0.3, 0.5] & \textcolor{red!80}{\ding{55}} & \textcolor{red!80}{\ding{55}} & 35.9 & 61.7 & 33.5 & 23.0 \\
                            & [0.3, 0.5] & \textcolor{highgreen}{\ding{51}} & \textcolor{red!80}{\ding{55}} & 36.8 & 63.2 & 34.3 & 24.0 \\
                            & [0.3, 0.5] & \textcolor{red!80}{\ding{55}} & \textcolor{highgreen}{\ding{51}} & 36.7 & 63.0 & 34.5 & 23.8 \\
                            & [0.3, 0.5] & \textcolor{highgreen}{\ding{51}} & \textcolor{highgreen}{\ding{51}} & 38.6 & 64.4 & 36.0 & 25.1 \\
        \hline\hline
        \end{tabular}}
        \midsepdefault
        \vspace{-0.2cm}
        \captionof{table}{\textbf{Ablation analysis of key components}. We present the ablation studies of CLIP-PGS's components, covering zero-shot image classification, linear probing, and text/image retrieval tasks.}
        \label{tab:6_ablation}
    \end{minipage}
    \hfill
    \begin{minipage}{0.25\linewidth}
        \centering
        \includegraphics[width=1.0\linewidth]{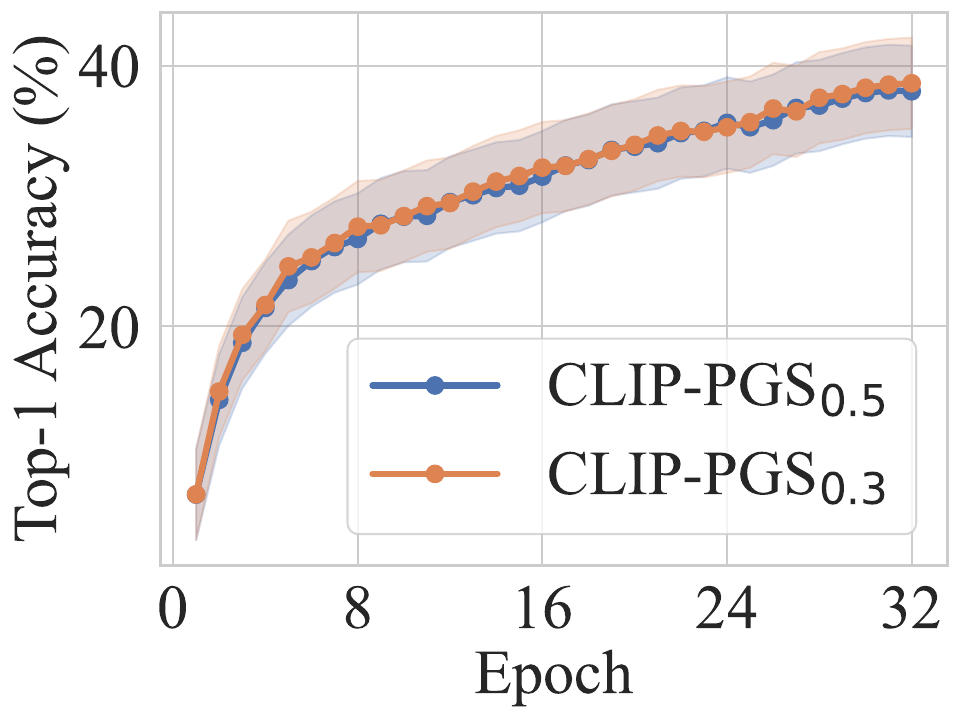}
        \vspace{-0.8cm}
        \caption{\textbf{Zero-shot classification on ImageNet-1K}~\cite{deng2009imagenet}. We present plots showing the trend of zero-shot accuracy across training epochs for the models trained on CC12M~\cite{cc12m} over 32 epochs.}
        \label{fig:5_acc}
    \end{minipage}
    \vspace{-0.2cm}
\end{figure*}

\noindent\textbf{Qualitative results.}
\cref{fig:4_patch_vis} illustrates the effectiveness of CLIP-PGS in selectively masking image regions while retaining critical semantic content.
By carefully masking non-essential areas, CLIP-PGS maintains the integrity of key elements, allowing for accurate descriptions without compromising important contextual details.
\cref{fig:5_acc} presents the progression of zero-shot top-1 accuracy on ImageNet-1K~\cite{deng2009imagenet} across training epochs for our models, CLIP-PGS$_{0.5}$ and CLIP-PGS$_{0.3}$, trained on CC12M~\cite{cc12m}.

\subsection{Ablation Studies} \label{sec:ablation}
This section provides a detailed ablation analysis of the core design components of CLIP-PGS. By default, we use ViT-B/16 as the image encoder, trained on CC12M~\cite{cc12m} for 32 epochs with a batch size of 4,096. \cref{tab:6_ablation} summarizes results on various downstream tasks, including zero-shot classification (ZS), linear probing (LP), and zero-shot text/image retrieval (TR/IR) on ImageNet-1K~\cite{deng2009imagenet} and MS-COCO~\cite{ms-coco}. Please refer to the appendix for more results.

\begin{figure}[t]
    \begin{center}
        \begin{subfigure}{0.49\linewidth}
            \centering
            \includegraphics[width=\linewidth]{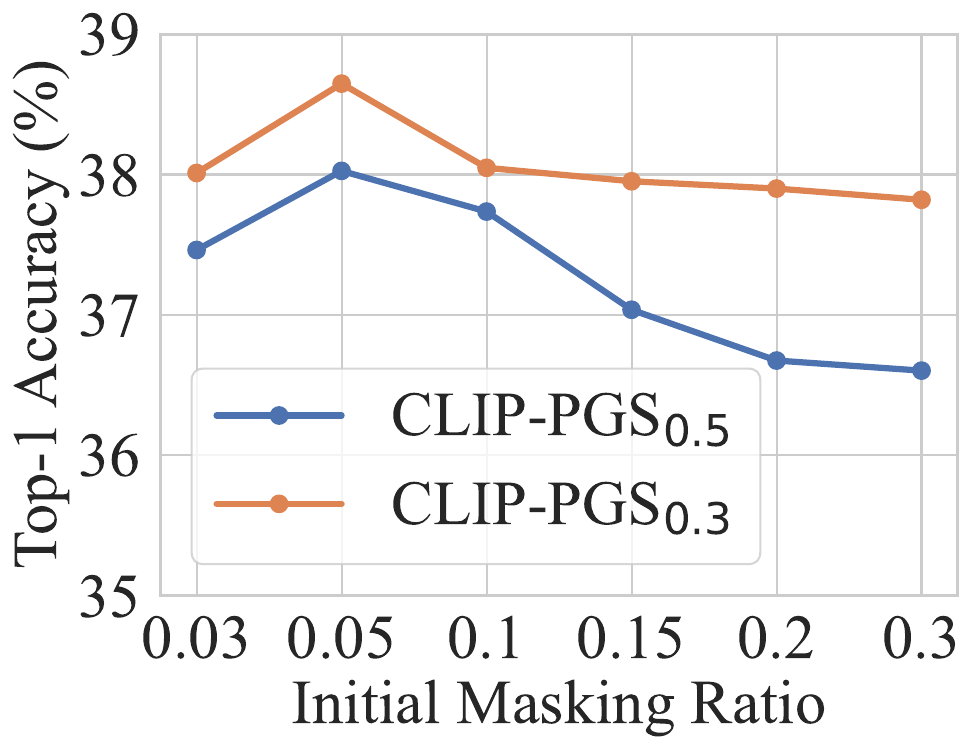}
            \caption{Top-1 \textit{vs.} Initial Masking Ratio}
            \label{fig6_a}
        \end{subfigure}
        \hfill
        \begin{subfigure}{0.49\linewidth}
            \centering
            \includegraphics[width=\linewidth]{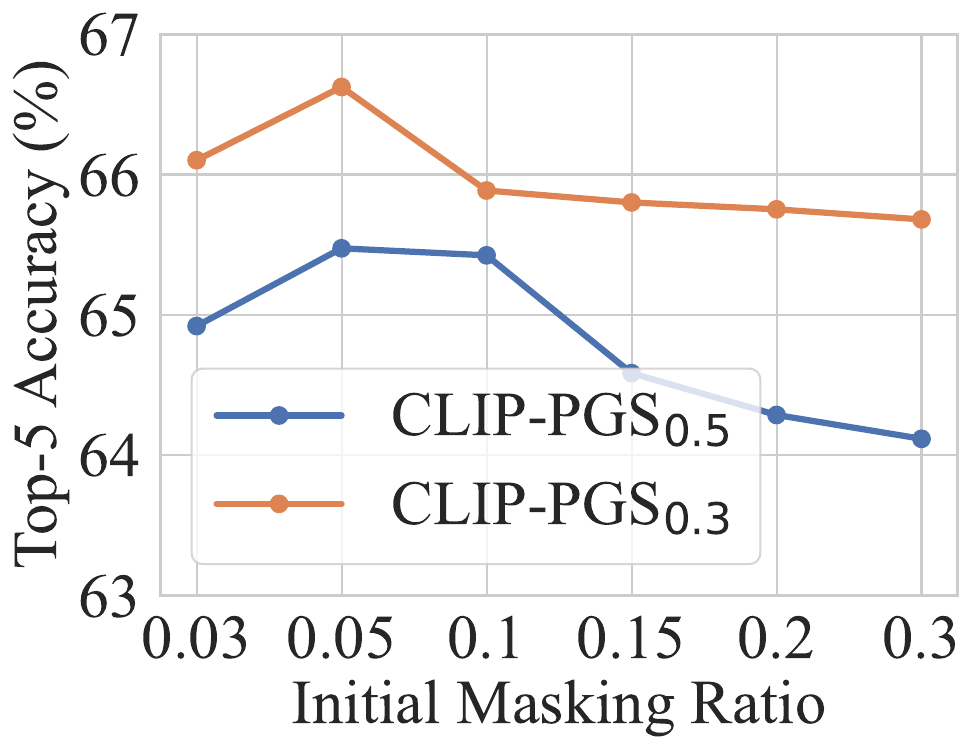}
            \caption{Top-5 \textit{vs.} Initial Masking Ratio}
            \label{fig6_b}
        \end{subfigure}
    \end{center}
    \vspace{-0.6cm}
    \caption{\textbf{Ablation analysis of initial masking ratio}. We report the zero-shot accuracy results of CLIP-PGS on ImageNet-1K~\cite{deng2009imagenet} at various initial masking ratios.}
    \label{fig:6_init_masking}
    \vspace{-0.35cm}
\end{figure}

\noindent\textbf{Initial Masking Ratio.}
We conduct ablations to assess the impact of varying initial masking ratios on CLIP-PGS, as shown in \cref{fig:6_init_masking}.
For top-1 accuracy, CLIP-PGS$_{0.3}$ reaches peak performance at lower initial masking ratios, particularly around 0.05. As the masking ratio increases, accuracy declines for both variants, though CLIP-PGS$_{0.3}$ maintains more stable results than CLIP-PGS$_{0.5}$. A similar trend appears at top-5, as higher masking ratios obscure key semantic information.
A lower initial masking ratio provides an optimal balance, preserving essential visual details while minimizing the risk of losing important semantic content.

\noindent\textbf{Lower/Upper Limit Masking Ratio} (\colorbox{red!25}{MR}).
Table \ref{tab:6_ablation} analyzes the influence of lower and upper masking limits on CLIP-PGS performance. Consistent with \cite{flip}, we set the upper masking ratio to 0.5, with subscripts in CLIP-PGS$_{0.5}$ and CLIP-PGS$_{0.3}$ indicating lower limits of 0.5 and 0.3, respectively. Thus, CLIP-PGS$_{0.5}$ employs a fixed masking ratio, while CLIP-PGS$_{0.3}$ dynamically adjusts between 0.3 and 0.5.
In contrast to FLIP's random masking, our approach utilizes progressive masking, starting with a small selection of mask patches and gradually expanding based on similarity scores. This method leads to improvements in zero-shot accuracy for both CLIP-PGS$_{0.5}$ and CLIP-PGS$_{0.3}$, achieving gains of \textbf{0.8\%} and \textbf{1.5\%}, respectively, and yielding comparable advancements in linear probing and zero-shot retrieval tasks. In particular, CLIP-PGS$_{0.3}$ consistently outperforms, with zero-shot classification accuracy increasing from 35.2\% (for CLIP-PGS$_{0.5}$) to 35.9\% without additional components, and reaching \textbf{38.6\%} with the addition of ED and OTN. A similar improvement appears in linear probing, where CLIP-PGS$_{0.3}$ attains \textbf{64.4\%}, surpassing the fixed 0.5 masking setup of CLIP-PGS$_{0.5}$.
Expanding the masking ratio range adds flexibility in preserving critical semantic information, thereby reinforcing CLIP-PGS's robustness and overall effectiveness.

\noindent\textbf{Edge Detection} (\colorbox{green!25}{ED}).
\cref{tab:6_ablation} presents the ablation results to analyze the effect of ED on CLIP-PGS. Adding ED yields consistent performance gains for both CLIP-PGS\(_{0.5}\) and CLIP-PGS\(_{0.3}\). For instance, CLIP-PGS\(_{0.3}\) increases zero-shot accuracy from 35.9\% to \textbf{36.8\%} and linear probing accuracy from 61.7\% to \textbf{63.2\%}. Text and image retrieval also benefit, underscoring ED's role in preserving key semantics and enhancing feature alignment within masked images. Edge detection is thus essential for improving model performance by emphasizing critical visual regions.

\noindent\textbf{Optimal Transport Normalization} (\colorbox{cyan!25}{OTN}).
\cref{tab:6_ablation} illustrates the impact of incorporating OTN on improving CLIP-PGS performance across all metrics. By establishing a balanced similarity matrix that prioritizes patches with higher similarity to adjacent regions, OTN preserves key semantic details, enhancing feature alignment in masked areas. 
For example, CLIP-PGS\(_{0.3}\) with OTN raises zero-shot accuracy from 35.9\% to \textbf{36.7\%} and linear probing from 61.7\% to \textbf{63.0\%}, with similar gains in text (\textbf{34.5\%}) and image retrieval (\textbf{23.8\%}).
Combined with ED, OTN yields the highest scores in zero-shot classification and retrieval tasks.

\noindent\textbf{Computational Efficiency}.
As shown in \cref{tab:6_ablation}, the additional computational cost of CLIP-PGS is minimal. Its key components contribute only a small overhead: MR (random masking and similarity computation) adds less than \textbf{1.0\%}, while ED (Sobel) and OTN (Sinkhorn) each contribute approximately \textbf{1\%}, resulting in a total overhead of less than \textbf{3\%}. Despite these additions, CLIP-PGS maintains training acceleration comparable to FLIP~\cite{flip}, ensuring high efficiency without compromising performance.

\begin{figure}[t]
    \begin{center}
        \begin{subfigure}{0.49\linewidth}
            \centering
            \includegraphics[width=\linewidth]{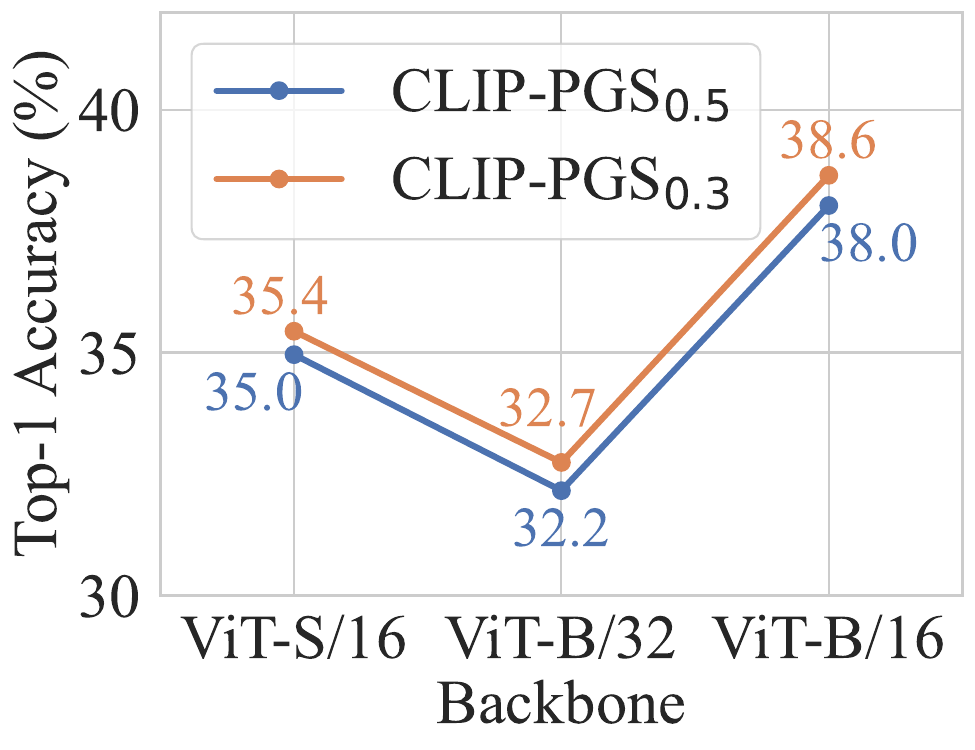}
            \caption{Top-1 \textit{vs.} Backbone}
            \label{fig7_a}
        \end{subfigure}
        \hfill
        \begin{subfigure}{0.49\linewidth}
            \centering
            \includegraphics[width=\linewidth]{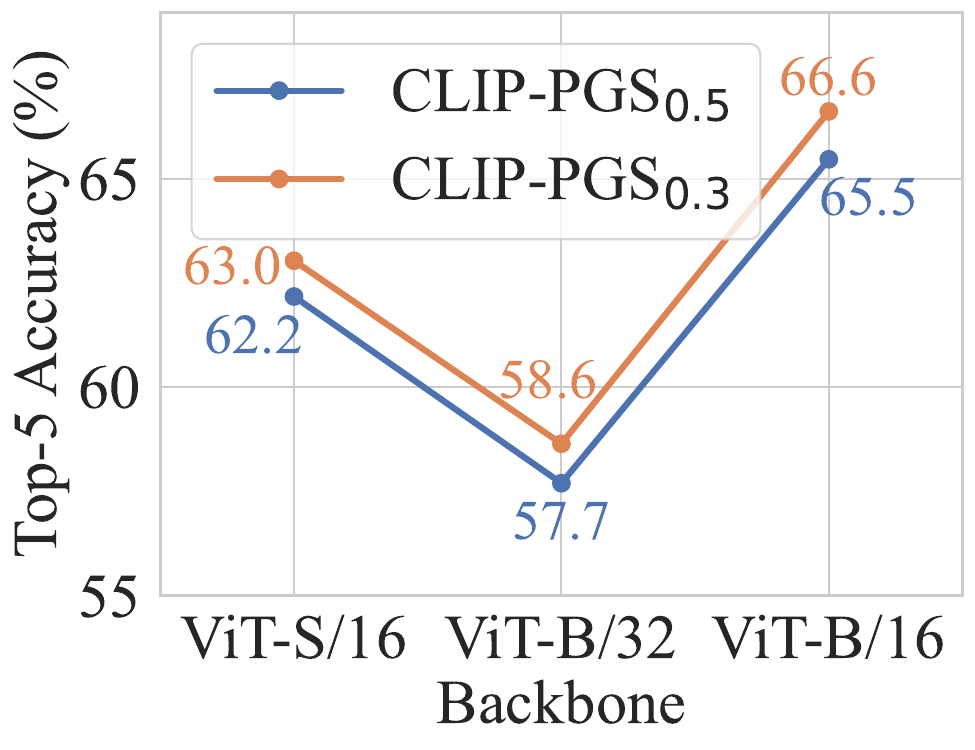}
            \caption{Top-5 \textit{vs.} Backbone}
            \label{fig7_b}
        \end{subfigure}
    \end{center}
    \vspace{-0.6cm}
    \caption{\textbf{Ablation analysis of different backbone and patch sizes}. We report the zero-shot accuracy results of CLIP-PGS on ImageNet-1K~\cite{deng2009imagenet} at various backbone sizes.}
    \label{fig:7_backbone}
    \vspace{-0.35cm}
\end{figure}

\noindent\textbf{Backbone and Patch Sizes.}
\cref{fig:7_backbone} provides the ablation results of different backbone and patch sizes. We assess three ViT~\cite{vit} configurations such as ViT-S/16, ViT-B/32 and ViT-B/16.
The ViT-B/16 backbone delivers the highest accuracy, demonstrating superior capability in capturing fine-grained features compared to the smaller ViT-S/16 and lower-resolution ViT-B/32. Our model, CLIP-PGS$_{0.3}$ with ViT-B/16, achieves \textbf{38.6\%} top-1 and \textbf{66.6\%} top-5 accuracy. 
The ablation results reinforce the widely accepted view that larger backbones with higher feature resolutions enhance the performance of vision-language pre-training models.

\section{Conclusion}
In this study, we introduce CLIP-PGS, a simple yet efficient masking framework for enhancing CLIP’s efficiency in vision-language alignment through Patch Generation-to-Selection. Our gradual approach begins by pre-selecting candidate patches for masking, followed by Sobel edge detection to create an edge mask that preserves key object regions. We further refine patch selection by computing similarity scores between candidate and neighboring patches, using optimal transport normalization to maintain balanced patch representation. This structured approach ensures effective retention of semantic information while reducing computational demands. CLIP-PGS achieves state-of-the-art performance in zero-shot classification and retrieval tasks. It also demonstrates significant improvements in robustness and language compositionality, showcasing its versatility across diverse downstream applications.

\noindent\textbf{Limitation and Future Work.}
Due to limited computational resources, our models are pre-trained only on the CC12M dataset, with primary experiments conducted on ViT-B/16. In future work, we aim to extend CLIP-PGS to convolutional network architectures~\cite{resnet,convnext}, broadening its applicability beyond the current transformer-based models. Additionally, while CLIP-PGS is designed for dual-encoder vision-language models like CLIP~\cite{clip}, our masking strategy is not inherently limited to this framework. We plan to explore its adaptation to other self-supervised learning approaches, such as masked image modeling techniques like MAE~\cite{he2022mae}, to potentially enhance their efficiency.

\small{\noindent\textbf{Acknowledgement.}
This work was supported by the National Natural Science Foundation of China (No. 62472222, 62222207, 62427808), Natural Science Foundation of Jiangsu Province (No. BK20240080).}

{
    \small
    \bibliographystyle{ieeenat_fullname}
    \bibliography{main}
}

\clearpage
\setcounter{page}{1}
\maketitlesupplementary

\appendix
\renewcommand\thefigure{\Alph{section}\arabic{figure}}
\renewcommand\thetable{\Alph{section}\arabic{table}}

\section*{Appendix} \label{sec:appendix}
This appendix presents implementation details (\S\ref{sec:imp_details}), supplementary experiments (\S\ref{sec:add_exps}), and additional qualitative results (\S\ref{sec:add_qual}).

\section{Implementation Details}\label{sec:imp_details}
\subsection{Architecture}
Consistent with the CLIP~\cite{clip} and OpenCLIP~\cite{openclip} frameworks, we utilize the ViT-B/16~\cite{vit} architecture as the primary image encoder backbone, paired with a 12-layer transformer featuring 512-dimensional embeddings and 8 attention heads for the text encoder. This appendix also includes experiments with ViT-S/16, a smaller vision backbone with 384-dimensional embeddings, 12 layers, and 6 attention heads. Table~\ref{tab:clip_model_configs} summarizes the detailed configurations of both ViT-S/16 and ViT-B/16, including their parameter counts and architectural specifications for both the vision and text encoders.

\subsection{Pre-training}
We provide the detailed pre-training settings for CLIP-PGS in Table~\ref{tab:pretraining_hyperparams}. The model is optimized using AdamW~\cite{adamw} with momentum parameters set to \((0.9, 0.98)\), a learning rate of \(1 \times 10^{-3}\), and a weight decay of 0.2. A cosine decay schedule~\cite{cosine_decay} with 10,000 warm-up steps is applied. The training is conducted for 32 epochs with a batch size of 4,096, utilizing an environment of 8 NVIDIA V100 GPUs (32G). These settings ensure efficient optimization and scalability during pre-training.

\begin{table}[b]
\vspace{-0.3cm}
\centering
\midsepremove
\setlength{\tabcolsep}{1.6pt}
\resizebox{1.0\linewidth}{!}{
\begin{tabular}{lcccccccccc}
\hline\hline
\multirow{2}{*}{\begin{tabular}[c]{@{}c@{}}Model\end{tabular}} & Embed & \multicolumn{3}{c}{Vision Transformer} & \multicolumn{3}{c}{Text Transformer} & \multicolumn{3}{c}{\# Params (M)} \\
\cline{3-5} \cline{6-8} \cline{9-11}
& Dim. & Layers & Width & Heads & Layers & Width & Heads & Vision & Text & Total \\
\midrule
S/16 & 384  & 12 & 384  & 6  & 12 & 384  & 6  & 22   & 33   & 55   \\
B/16 & 512  & 12 & 768  & 12 & 12 & 512  & 8  & 86   & 53   & 141  \\
\hline\hline
\end{tabular}}
\midsepdefault
\vspace{-0.2cm}
\caption{\textbf{Detailed configuration of encoder architectures}, including embedding dimensions (Dim.), transformer specifications for vision and text encoders, and parameter counts.}
\label{tab:clip_model_configs}
\vspace{-0.2cm}
\end{table}

\begin{table}[b]
\centering
\midsepremove
\setlength{\tabcolsep}{19.6pt}
\resizebox{0.95\linewidth}{!}{
\begin{tabular}{ll}
\hline\hline
Config            & Value                \\
\midrule
optimizer                  & AdamW~\cite{adamw}                  \\
optimizer momentum         & (0.9, 0.98)                         \\
batch size                 & 4,096                               \\
learning rate              & 1e-3                                \\
warm-up steps              & 10,000                              \\
schedule                   & cosine decay~\cite{cosine_decay}    \\
weight decay               & 0.2                                 \\
training epochs            & 32                                  \\
GPU environment            & 8 NVIDIA V100 GPUs (32G)            \\
\hline\hline
\end{tabular}}
\midsepdefault
\vspace{-0.2cm}
\caption{Pre-training settings of CLIP-PGS.}
\label{tab:pretraining_hyperparams}
\vspace{-0.2cm}
\end{table}

\section{Supplementary Experiments}\label{sec:add_exps}
\subsection{Downstream Datasets.}
\cref{tab:dataset_overview} provides a comprehensive overview of the datasets utilized in our experiments, detailing the number of classes, training and testing set sizes, and their corresponding evaluation tasks (\eg, recognition, robustness, and retrieval).

\begin{table}[t]
\centering
\midsepremove
\setlength{\tabcolsep}{1.0pt}
\resizebox{1.0\linewidth}{!}{
\begin{tabular}{lcccc}
\hline\hline
Dataset     & Classes & Train Size & Test Size &  Evaluation Task \\
\midrule
Food101~\cite{food101}       & 101  & 75,750  & 25,250  & fine-grained recognition \\
CIFAR10~\cite{cifar}         & 10   & 45,000  & 10,000  & fine-grained recognition \\
CIFAR100~\cite{cifar}        & 100  & 45,000  & 10,000  & fine-grained recognition \\
SUN397~\cite{sun397}         & 397  & -       & 108,754 & scene recognition\\
Cars~\cite{cars}             & 196  & 8,144   & 8,041   & fine-grained recognition \\
VOC2007~\cite{voc2007}       & 20   & 7,844   & 14,976  & object recognition \\
Aircraft~\cite{aircraft}     & 100  & 3,334   & 3,333   & fine-grained recognition\\
DTD~\cite{dtd}               & 47   & 1,880   & 1,880   & texture recognition \\
OxfordPets~\cite{pets}       & 37   & 2,944   & 3,669   & fine-grained recognition \\
Caltech101~\cite{caltech101} & 102  & 2,753   & 6,085   & object recognition \\
Flowers~\cite{flowers}       & 102  & 1,020   & 6,149   & fine-grained recognition \\
STL10~\cite{stl10}           & 10   & 5,000   & 8,000   & object recognition \\
EuroSAT~\cite{eurosat}       & 10   & 16,200  & 5,400   & aerial image recognition \\
RESISC45~\cite{resisc45}     & 45   & 18,900  & 6,300   & aerial image recognition \\
GTSRB~\cite{gtsrb}           & 43   & 26,640  & 12,630  & traffic sign recognition \\
Country211~\cite{yfcc100m}   & 211  & 31,650  & 21,100  & geo-tagged recognition \\
PCam~\cite{pcam}             & 2    & 262,144 & 32,768  & digital pathology \\
\midrule
ImageNet-1K~\cite{deng2009imagenet}    & 1000 & 1,281,167 & 50,000 & fine-grained recognition \\
ImageNet-V2~\cite{imagenet-v2}         & 1000 & -         & 10,000 & robustness of collocation \\
ImageNet-A~\cite{imagenet-ao}          & 200  & -         & 7,500  & robustness of attack \\
ImageNet-R~\cite{imagenet-r}           & 200  & -         & 30,000 & robustness of multi-domains \\
ImageNet-O~\cite{imagenet-ao}          & 200  & -         & 7,500  & robustness of attack \\
ImageNet-Sketch~\cite{imagenet-sketch} & 1000 & -         & 50,889 & robustness of sketch domain \\
\midrule
MS-COCO~\cite{ms-coco}   & - & 82,783 & 5,000 & text/image retrieval \\
Flickr8K~\cite{flickr}   & - & 6,000  & 1,000 & text/image retrieval \\
Flickr30K~\cite{flickr}  & - & 29,000 & 1,000 & text/image retrieval \\
\hline\hline
\end{tabular}}
\midsepdefault
\vspace{-0.2cm}
\caption{\textbf{Overview of downstream datasets}, including the number of classes, training and testing set sizes, and evaluation tasks.}
\label{tab:dataset_overview}
\vspace{-0.2cm}
\end{table}

\subsection{Downstream Evaluation Tasks}
In this supplementary material, we expand the evaluation to include results with ViT-S/16~\cite{vit} as the visual backbone, complementing the main text. We assess the model across \textbf{five} standard benchmark scenarios: zero-shot classification, zero-shot text/image retrieval, linear probing, robustness evaluation, and language compositionality, adhering to established evaluation protocols~\cite{clip,slip,openclip,flip}.
\textsuperscript{\ref{clip-benchmark}}

\begin{itemize}
\setlength{\itemsep}{2.3pt}
\setlength{\parsep}{0pt}
\setlength{\parskip}{0pt}

    \item \textbf{Zero-shot classification} (\cref{tab:appendix_zsc}): We evaluate model generalizability on 17 datasets, including Food101~\cite{food101}, CIFAR10~\cite{cifar}, and ImageNet-1K~\cite{deng2009imagenet}. These benchmarks assess performance under varying distributional shifts, highlighting the robustness of our approach.

    \item \textbf{Zero-shot retrieval} (\cref{tab:appendix_zsr}): Text-to-image and image-to-text retrieval tasks are conducted on MS-COCO~\cite{ms-coco} and Flickr~\cite{flickr}. These benchmarks evaluate the model’s capability to associate visual and language representations without additional fine-tuning.

    \item \textbf{Linear probing} (\cref{tab:appendix_lp}): Visual representations are assessed on ImageNet-1K~\cite{deng2009imagenet}, CIFAR10~\cite{cifar}, and CIFAR100~\cite{cifar} using linear classifiers. Following the clip-benchmark setup \textsuperscript{\ref{clip-benchmark}}, we train for 10 epochs with AdamW optimizer~\cite{adamw}, a 0.1 learning rate, and a batch size of 64.

    \item \textbf{Robustness evaluation} (\cref{tab:appendix_robust}): The model is tested on ImageNet-1K~\cite{deng2009imagenet} and out-of-distribution datasets such as ImageNet-V2~\cite{imagenet-v2}, ImageNet-A~\cite{imagenet-ao}, and ImageNet-Sketch~\cite{imagenet-sketch}. This evaluation examines resilience to distributional shifts.

    \item \textbf{Language compositionality} (\cref{tab:appendix_composition}): Performance on the SugarCrepe~\cite{sugarcrepe} dataset is used to assess adaptability to complex language structures, including manipulations of objects, attributes, and relations, showcasing the model's precision in aligning visual and linguistic cues.
\end{itemize}

\begin{table*}[t]
\centering
\midsepremove
\setlength{\tabcolsep}{3.0pt}
\resizebox{1.0\linewidth}{!}{
\begin{tabular}{lc||ccccccccccccccccc||c}
\hline\hline
Method & Image Enc. & \rotatebox{65}{Food101} & \rotatebox{65}{CIFAR10} & \rotatebox{65}{CIFAR100} & \rotatebox{65}{SUN397} & \rotatebox{65}{Cars} & \rotatebox{65}{VOC2007} & \rotatebox{65}{Aircraft} & \rotatebox{65}{DTD} & \rotatebox{65}{OxfordPets} & \rotatebox{65}{Caltech101} & \rotatebox{65}{Flowers} & \rotatebox{65}{STL10} & \rotatebox{65}{EuroSAT} & \rotatebox{65}{RESISC45} & \rotatebox{65}{GTSRB} & \rotatebox{65}{Country211} & \rotatebox{65}{PCam} & \rotatebox{65}{Average} \\
\midrule
CLIP~\cite{clip} & ViT-B/16   & 42.3 & 57.7 & 25.0 & 44.1 & 17.0 & 50.5 & 1.7 & 16.5 & 53.9 & \underline{73.5} & 26.0 & 82.0 & 18.7 & 26.5 & 9.4 & \underline{4.5} & 48.0 & 35.1 \\ 
FLIP~\cite{flip} & ViT-B/16   & 39.9 & 52.8 & 24.5 & 42.8 & 15.9 & 46.6 & 1.4 & 15.9 & 46.0 & 70.4 & 25.3 & 80.2 & 17.0 & 25.8 & 5.6 & 4.0 & 47.1 & 33.0 \\ 
A-CLIP~\cite{a-clip} & ViT-B/16 & 41.8 & 61.6 & 27.1 & \underline{46.6} & 16.0 & \underline{51.1} & 1.3 & 17.1 & 51.2 & \underline{73.5} & 25.7 & 85.8 & 20.5 & 29.1 & 8.0 & 4.2 & 50.1 & 35.9 \\ 
E-CLIP~\cite{e-clip} & ViT-B/16 & 42.1 & \underline{70.7} & 32.0 & 43.9 & 15.1 & 43.6 & 2.2 & 17.0 & 55.4 & \textbf{73.7} & 28.4 & 85.6 & \underline{22.9} & 30.0 & 9.6 & \textbf{4.7} & 50.0 & 36.9 \\ 
\midrule
\rowcolor{rowgray} \textit{Ours} &&&&&&&&&&&&&&&&&&&  \\ 
\rowcolor{rowblue} \textbf{CLIP-PGS}$_{0.5}$ & ViT-B/16 & \underline{42.8} & 62.5 & \underline{35.5} & 45.5 & \underline{17.3} & 50.0 & 1.9 & 17.4 & \underline{55.7} & 71.8 & \textbf{33.2} & \underline{88.2} & 20.5 & \textbf{31.8} & 10.1 & \textbf{4.7} & 50.0 & \underline{37.6} \\ 
\rowcolor{rowblue} \textbf{CLIP-PGS}$_{0.3}$ & ViT-B/16 & \textbf{46.5} & \textbf{73.5} & \textbf{37.3} & \textbf{47.5} & \textbf{19.9} & \textbf{55.1} & \textbf{3.1} & \textbf{19.8} & \textbf{58.1} & 72.7 & \underline{30.7} & \underline{88.2} & 22.8 & \underline{30.4} & \underline{10.9} & \underline{4.5} & \underline{50.8} & \textbf{39.5} \\ 
\midrule
\rowcolor{rowgreen} \textbf{CLIP-PGS}$_{0.5}$ & ViT-S/16 & 38.7 & 58.4 & 29.0 & 43.7 & 12.7 & 48.0 & 2.0 & \underline{17.7} & 50.6 & 69.4 & 26.6 & 86.4 & \textbf{27.6} & 25.9 & \textbf{11.2} & 3.9 & 56.4 & 35.8\\ 
\rowcolor{rowgreen} \textbf{CLIP-PGS}$_{0.3}$ & ViT-S/16 & 39.1 & 66.9 & 30.8 & 44.0 & 15.0 & 47.1 & \underline{2.7} & 14.8 & 54.5 & 71.8 & 28.5 & \textbf{88.3} & 16.6 & 25.6 & 7.9 & 4.3 & 50.2 & 35.8 \\ 
\hline\hline
\end{tabular}}
\midsepdefault
\vspace{-0.2cm}
\caption{\textbf{Zero-shot classification results}. We evaluate performance on 17 diverse classification datasets, reporting both top-1 accuracy (\%) and the overall average. The optimal result is highlighted in \textbf{bold}, and the second-best result is \underline{underlined}.}
\label{tab:appendix_zsc}
\vspace{-0.2cm}
\end{table*}

\begin{table*}[t]
\centering
\midsepremove
\setlength{\tabcolsep}{3.2pt}
\resizebox{1.0\linewidth}{!}{
\begin{tabular}{lc||ccccccccc||ccccccccc}
\hline\hline
& & \multicolumn{9}{c||}{Text Retrieval} & \multicolumn{9}{c}{Image Retrieval} \\
\cmidrule(lr){3-11} \cmidrule(lr){12-20}
Method & Image Enc. & \multicolumn{3}{c}{MS-COCO} & \multicolumn{3}{c}{Flickr8K} & \multicolumn{3}{c||}{Flickr30K} & \multicolumn{3}{c}{MS-COCO} & \multicolumn{3}{c}{Flickr8K} & \multicolumn{3}{c}{Flickr30K}\\
\cmidrule(lr){3-5} \cmidrule(lr){6-8} \cmidrule(lr){9-11} \cmidrule(lr){12-14} \cmidrule(lr){15-17} \cmidrule(lr){18-20}
& & R@1 & R@5 & R@10 & R@1 & R@5 & R@10 & R@1 & R@5 & R@10 & R@1 & R@5 & R@10 & R@1 & R@5 & R@10 & R@1 & R@5 & R@10 \\
\midrule
CLIP~\cite{clip} & ViT-B/16 & 34.6 & \underline{62.0} & 72.7 & 55.7 & 81.6 & 89.9 & \underline{58.5} & \underline{83.8} & 89.1 & 23.5 & 47.8 & 59.7 & 40.5 & 68.9 & 80.2 & 43.2 & 70.4 & 80.4 \\
FLIP~\cite{flip} & ViT-B/16 & 32.6 & 59.1 & 70.6 & 55.0 & 80.9 & 88.9 & 53.8 & 80.8 & 88.5 & 22.6 & 46.1 & 58.1 & 40.3 & 68.1 & 78.6 & 41.5 & 67.9 & 77.5 \\
A-CLIP~\cite{a-clip} & ViT-B/16 & 33.7 & 60.2 & 71.0 & 53.7 & 80.1 & 88.0 & 55.3 & 81.4 & 87.6 & 23.9 & 48.3 & 60.0 & 40.6 & 68.9 & 78.9 & 43.1 & 70.1 & 78.8 \\
E-CLIP~\cite{e-clip} & ViT-B/16 & 34.3 & \underline{62.0} & \underline{73.3} & 57.0 & 82.7 & 90.1 & 55.8 & \textbf{84.2} & 89.6 & 23.8 & 48.2 & 59.8 & 42.0 & 69.4 & 79.6 & 43.3 & 70.9 & 80.2 \\
\midrule
\rowcolor{rowgray} \textit{Ours} &&&&&&&&&&&&&&&&&&&  \\ 
\rowcolor{rowblue} \textbf{CLIP-PGS}$_{0.5}$ & ViT-B/16 & \underline{35.2} & 61.9 & 72.8 & \textbf{58.5} & \textbf{83.6} & \underline{90.6} & 57.7 & 82.7 & \underline{90.4} & \underline{24.3} & \underline{48.8} & \underline{60.5} & \underline{43.5} & \underline{70.7} & \underline{81.0} & \underline{45.3} & \underline{72.9} & \underline{81.2} \\
\rowcolor{rowblue} \textbf{CLIP-PGS}$_{0.3}$ & ViT-B/16 & \textbf{36.0} & \textbf{64.4} & \textbf{74.6} & \underline{58.3} & \underline{82.9} & \textbf{90.8} & \textbf{59.9} & 83.5 & \textbf{90.8} & \textbf{25.1} & \textbf{49.5} & \textbf{61.6} & \textbf{44.4} & \textbf{71.7} & \textbf{81.1} & \textbf{47.1} & \textbf{73.5} & \textbf{82.0} \\
\midrule
\rowcolor{rowgreen} \textbf{CLIP-PGS}$_{0.5}$ & ViT-S/16 & 31.6 & 58.1 & 69.6 & 53.6 & 81.5 & 89.3 & 53.9 & 80.0 & 87.4 & 22.2 & 45.6 & 57.6 & 39.8 & 68.1 & 78.8 & 41.4 & 68.2 & 77.3 \\
\rowcolor{rowgreen} \textbf{CLIP-PGS}$_{0.3}$ & ViT-S/16 & 33.1 & 60.5 & 72.4 & 52.3 & 80.9 & 89.7 & 55.9 & 80.5 & 88.1 & 22.9 & 46.7 & 58.6 & 40.2 & 68.1 & 79.4 & 42.7 & 69.1 & 78.0 \\
\hline\hline
\end{tabular}}
\midsepdefault
\vspace{-0.2cm}
\caption{\textbf{Zero-shot text/image retrieval results}. We evaluate performance on the MS-COCO~\cite{ms-coco}, Flickr8k~\cite{flickr}, and Flickr30k~\cite{flickr} datasets, reporting Recall@1 (\%, R@1), Recall@5 (\%, R@5), and Recall@10 (\%, R@10) for both text and image retrieval tasks.}
\label{tab:appendix_zsr}
\vspace{-0.2cm}
\end{table*}

\begin{table}[t]
\vspace{0.2cm}
\centering
\midsepremove
\setlength{\tabcolsep}{3.5pt}
\resizebox{1.0\linewidth}{!}{
\begin{tabular}{lc||ccc}
\hline\hline
Method & Image Enc. & \rotatebox{0}{CIFAR10} & \rotatebox{0}{CIFAR100} & \rotatebox{0}{ImageNet-1K} \\ 
\midrule
CLIP~\cite{clip}     & ViT-B/16 & 88.0 & 67.4 & 62.3  \\ 
FLIP~\cite{flip}     & ViT-B/16 & 85.9 & 65.5 & 61.3  \\ 
A-CLIP~\cite{a-clip} & ViT-B/16 & 86.4 & 66.1 & 62.0 \\ 
E-CLIP~\cite{e-clip} & ViT-B/16 & 89.0 & 69.7 & 62.7 \\ 
\midrule
\rowcolor{rowgray} \textit{Ours} &&&& \\
\rowcolor{rowblue} \textbf{CLIP-PGS}$_{0.5}$ & ViT-B/16 & \underline{89.5} \textcolor{highgreen}{\textbf{(+0.5)}} & \underline{70.3} \textcolor{highgreen}{\textbf{(+0.6)}} & \underline{64.2} \textcolor{highgreen}{\textbf{(+1.5)}} \\
\rowcolor{rowblue} \textbf{CLIP-PGS}$_{0.3}$ & ViT-B/16 & \textbf{90.0} \textcolor{highgreen}{\textbf{(+1.0)}} & \textbf{72.3} \textcolor{highgreen}{\textbf{(+2.6)}} & \textbf{64.4} \textcolor{highgreen}{\textbf{(+1.7)}} \\
\midrule
\rowcolor{rowgreen} \textbf{CLIP-PGS}$_{0.5}$ & ViT-S/16 & 87.7 & 68.6 & 62.7 \\
\rowcolor{rowgreen} \textbf{CLIP-PGS}$_{0.3}$ & ViT-S/16 & 88.1 & 68.7 & 62.9 \\
\hline\hline
\end{tabular}}
\midsepdefault
\vspace{-0.2cm}
\caption{\textbf{Linear probing classification results}. We evaluate all models on three common datasets, \ie, CIFAR10~\cite{cifar}, CIFAR100~\cite{cifar}, and ImageNet-1K~\cite{deng2009imagenet}, training each for 10 epochs under a consistent linear training setup. We present top-1 accuracy (\%), with gains over the stronger baseline highlighted in \textcolor{highgreen}{\textbf{(green)}}.}
\label{tab:appendix_lp}
\vspace{-0.2cm}
\end{table}

\begin{table*}[ht]
\centering
\midsepremove
\setlength{\tabcolsep}{2.8pt}
\resizebox{1.0\linewidth}{!}{
\begin{tabular}{lc||cccccc||ccc}
\hline\hline
Method & Image Enc. & \rotatebox{0}{ImageNet-1K} & \rotatebox{0}{ImageNet-V2} & \rotatebox{0}{ImageNet-A} & \rotatebox{0}{ImageNet-R} & \rotatebox{0}{ImageNet-O} & \rotatebox{0}{ImageNet-Sketch} & \rotatebox{0}{Average} & \rotatebox{0}{ID Average} & \rotatebox{0}{OOD Average} \\ 
\midrule
CLIP~\cite{clip}     &ViT-B/16 & 36.1 & 30.7 & 8.0 & 47.6 & 38.4 & 24.9 & 31.0 & 36.1 & 29.0 \\ 
FLIP~\cite{flip}     &ViT-B/16 & 34.4 & 29.5 & 7.1 & 41.4 & 39.5 & 20.1 & 28.7 & 34.4 & 27.5 \\ 
A-CLIP~\cite{a-clip} &ViT-B/16 & 35.2 & 30.1 & 8.1 & 45.1 & 39.4 & 23.7 & 30.3 & 35.2 & 30.3 \\ 
E-CLIP~\cite{e-clip} &ViT-B/16 & 36.3 & 30.7 & 8.1 & \underline{47.9} & 39.6 & \underline{25.4} & 31.3 & 36.3 & 30.3 \\ 
\midrule
\rowcolor{rowgray} \textit{Ours} &&&&&&&&&&  \\
\rowcolor{rowblue} \textbf{CLIP-PGS}$_{0.5}$ &ViT-B/16 & \underline{38.0} & \underline{32.6} & \underline{9.1} & 45.1 & \underline{41.1} & 23.9 & \underline{31.6} & \underline{38.0} & \underline{30.4} \\  
\rowcolor{rowblue} \textbf{CLIP-PGS}$_{0.3}$ &ViT-B/16 & \textbf{38.6} & \textbf{33.1} & \textbf{9.6} & \textbf{48.1} & \textbf{42.6} & \textbf{25.6} & \textbf{32.9} & \textbf{38.6} & \textbf{31.8} \\
\midrule
\rowcolor{rowgreen} \textbf{CLIP-PGS}$_{0.5}$ &ViT-S/16 & 34.9 & 29.5 & 7.4 & 41.8 & 40.6 & 21.4 & 29.3 & 34.9 & 28.1 \\  
\rowcolor{rowgreen} \textbf{CLIP-PGS}$_{0.3}$ &ViT-S/16 & 35.4 & 30.2 & 7.8 & 43.5 & 40.4 & 22.5 & 30.0 & 35.4 & 28.9 \\
\hline\hline
\end{tabular}}
\midsepdefault
\vspace{-0.2cm}
\caption{\textbf{Robustness assessment results}. We evaluate model robustness on ImageNet-1K~\cite{deng2009imagenet} and five of its variants~\cite{imagenet-v2,imagenet-ao,imagenet-r,imagenet-sketch}, reporting top-1 accuracy (\%) along with overall averages for in-distribution (ID) and out-of-distribution (OOD) performance.}
\label{tab:appendix_robust}
\vspace{-0.2cm}
\end{table*}

\begin{table*}[ht]
\centering
\midsepremove
\setlength{\tabcolsep}{6.8pt}
\resizebox{0.96\linewidth}{!}{
\begin{tabular}{lc||ccccccc||ccc}
\hline\hline
\multirow{2}{*}{\begin{tabular}[c]{@{}c@{}}Method\end{tabular}} & \multirow{2}{*}{\begin{tabular}[c]{@{}c@{}}Image Enc.\end{tabular}} &  \multicolumn{3}{c}{REPLACE} & \multicolumn{2}{c}{SWAP} & \multicolumn{2}{c||}{ADD} & \multicolumn{3}{c}{Average} \\
\cmidrule(lr){3-5} \cmidrule(lr){6-7} \cmidrule(lr){8-9} \cmidrule(lr){10-12}
& & Object & Attribute & Relation & Object & Attribute & Object & Attribute & Object & Attribute & Relation \\
\midrule
CLIP~\cite{clip} & ViT-B/16 & 85.8 & \textbf{79.2} & 64.5 & 61.8 & 58.7 & 74.2 & 68.4 & \underline{73.7} & \underline{68.8} & 64.5 \\
FLIP~\cite{flip} & ViT-B/16 & 84.1 & 75.9 & \underline{66.0} & 60.2 & 61.6 & 71.7 & 63.2 & 72.0 & 66.9 & \underline{66.0} \\
A-CLIP~\cite{a-clip} & ViT-B/16 & 86.6 & 75.5 & 63.2 & 52.4 & 63.1 & 71.6 & 66.8 & 71.6 & 68.4 & 63.2 \\
E-CLIP~\cite{e-clip} & ViT-B/16 & \underline{86.9} & 73.5 & 60.2 & 59.4 & 63.4 & 73.3 & 66.8 & 73.2 & 68.4 & 60.2 \\
\midrule
\rowcolor{rowgray} \textit{Ours} &&&&&&&&&&& \\
\rowcolor{rowblue} \textbf{CLIP-PGS}$_{0.5}$ & ViT-B/16 & 86.0 & 77.0 & 64.6 & \underline{63.3} & \underline{65.5} & \textbf{77.3} & \underline{69.8} & \textbf{75.5} & \textbf{70.8} & 64.6 \\
\rowcolor{rowblue} \textbf{CLIP-PGS}$_{0.3}$ & ViT-B/16 & \textbf{88.1} & 76.0 & \textbf{67.9} & \textbf{64.1} & \textbf{66.5} & 74.2 & \textbf{69.9} & \textbf{75.5} & \textbf{70.8} & \textbf{67.9} \\
\midrule
\rowcolor{rowgreen} \textbf{CLIP-PGS}$_{0.5}$ & ViT-S/16 & 84.9 & 76.5 & 65.4 & 60.4 & 65.0 & 73.6 & \textbf{69.9} & 73.0 & 70.5 & 65.4 \\
\rowcolor{rowgreen} \textbf{CLIP-PGS}$_{0.3}$ & ViT-S/16 & 86.6 & \underline{77.8} & 63.9 & 58.8 & 63.4 & \underline{74.9} & 68.4 & 73.4 & 70.2 & 63.9 \\
\hline\hline
\end{tabular}}
\midsepdefault
\vspace{-0.2cm}
\caption{\textbf{Language compositionality results}. We evaluate the compositionality of vision-language models on the SugarCrepe~\cite{sugarcrepe} dataset, which tests models by generating mismatched captions by replacing, swapping, or adding fine-grained atomic concepts (object, attribute, and relation). We report Recall@1 (\%) and the overall average for each atomic concept.}
\label{tab:appendix_composition}
\vspace{-0.5cm}
\end{table*}

\subsection{Ablation Studies}
This section presents an in-depth ablation analysis of the key design components in CLIP-PGS. Unless otherwise specified, experiments use ViT-B/16 as the image encoder, trained on the CC12M dataset~\cite{cc12m} for 32 epochs with a batch size of 4,096. We evaluate performance on diverse downstream tasks, including zero-shot classification (ZS), linear probing (LP), and zero-shot text/image retrieval (TR/IR), as summarized in \cref{tab:appendix_ablation}. For edge detection (ED), we compare the commonly used Sobel operator with the Canny edge detector to investigate their impact on performance.

\begin{table}[t]
\centering
\midsepremove
\setlength{\tabcolsep}{3.3pt}
\resizebox{1.0\linewidth}{!}{
\begin{tabular}{l||ccccccc}
\hline\hline
\multirow{2}{*}{Method} & \multicolumn{3}{c}{Component} & \multicolumn{2}{c}{ImageNet-1K} & \multicolumn{2}{c}{MS-COCO} \\
\cmidrule(lr){2-4} \cmidrule(lr){5-6} \cmidrule(lr){7-8}
& \colorbox{red!25}{MR} & \colorbox{green!25}{ED} & \colorbox{cyan!25}{OTN} & ZS & LP & TR & IR \\
\midrule
\rowcolor{rowgray} \textcolor{gray}{\textit{Baseline}} &&&&&&& \\
\rowcolor{rowgray} \textcolor{gray}{CLIP~\cite{clip}}
                    & \textcolor{gray}{-}     & \textcolor{gray}{-} & \textcolor{gray}{-} & \textcolor{gray}{36.1} & \textcolor{gray}{62.3} & \textcolor{gray}{34.6} & \textcolor{gray}{23.5} \\
\midrule
\rowcolor{rowgray} \textcolor{gray}{\textit{Random Mask}} &&&&&&& \\
\rowcolor{rowgray} \textcolor{gray}{FLIP~\cite{flip}}
                    & \textcolor{gray}{0.5}   & \textcolor{gray}{-} & \textcolor{gray}{-} & \textcolor{gray}{34.4} & \textcolor{gray}{61.3} & \textcolor{gray}{32.6} & \textcolor{gray}{22.6} \\
\midrule
\multirow{4}{*}{\textbf{CLIP-PGS}$_{0.5}$}
                    & 0.5        & \textcolor{red!80}{\ding{55}} & \textcolor{red!80}{\ding{55}} & 35.2 & 61.9 & 33.7 & 22.8 \\
                    & 0.5        & \textcolor{highgreen}{\ding{51}} & \textcolor{red!80}{\ding{55}} & 36.2 & 62.8 & 34.1 & 23.4 \\
                    & 0.5        & \textcolor{highgreen}{\ding{51}}$^*$ & \textcolor{red!80}{\ding{55}} & 35.8 & 62.7 & 34.0 & 23.2 \\
                    & 0.5        & \textcolor{red!80}{\ding{55}} & \textcolor{highgreen}{\ding{51}} & 36.3 & 62.7 & 33.9 & 23.2\\
                    & 0.5        & \textcolor{highgreen}{\ding{51}} & \textcolor{highgreen}{\ding{51}} & 38.0 & 64.2 & 35.2 & 24.3 \\
                    & 0.5        & \textcolor{highgreen}{\ding{51}}$^*$ & \textcolor{highgreen}{\ding{51}} & 37.8 & 64.1 & 35.1 & 24.0 \\
\midrule
\multirow{4}{*}{\textbf{CLIP-PGS}$_{0.3}$}
                    & [0.3, 0.5] & \textcolor{red!80}{\ding{55}} & \textcolor{red!80}{\ding{55}} & 35.9 & 61.7 & 33.5 & 23.0 \\
                    & [0.3, 0.5] & \textcolor{highgreen}{\ding{51}} & \textcolor{red!80}{\ding{55}} & 36.8 & 63.2 & 34.3 & 24.0 \\
                    & [0.3, 0.5] & \textcolor{highgreen}{\ding{51}}$^*$ & \textcolor{red!80}{\ding{55}} & 36.7 & 63.0 & 34.0 & 23.9 \\
                    & [0.3, 0.5] & \textcolor{red!80}{\ding{55}} & \textcolor{highgreen}{\ding{51}} & 36.7 & 63.0 & 34.5 & 23.8 \\
                    & [0.3, 0.5] & \textcolor{highgreen}{\ding{51}} & \textcolor{highgreen}{\ding{51}} & 38.6 & 64.4 & 36.0 & 25.1 \\
                    & [0.3, 0.5] & \textcolor{highgreen}{\ding{51}}$^*$ & \textcolor{highgreen}{\ding{51}} & 38.5 & 64.4 & 35.7 & 24.9 \\
\hline\hline
\end{tabular}}
\midsepdefault
\vspace{-0.2cm}
\caption{\textbf{Ablation analysis of key components}. We present comprehensive ablation experiments of CLIP-PGS's components, covering zero-shot image classification, linear probing, and text/image retrieval tasks. Here, \colorbox{red!25}{MR} stands for masking ratio, \colorbox{green!25}{ED} for edge detection, and \colorbox{cyan!25}{OTN} for optimal transport normalization. `*' denotes the use of the Canny edge detection method, Sobel is used by default.}
\label{tab:appendix_ablation}
\vspace{-0.2cm}
\end{table}

\section{Additional Qualitative Results}\label{sec:add_qual}
\begin{figure}[t]
    \centering
    \includegraphics[width=1.0\linewidth]{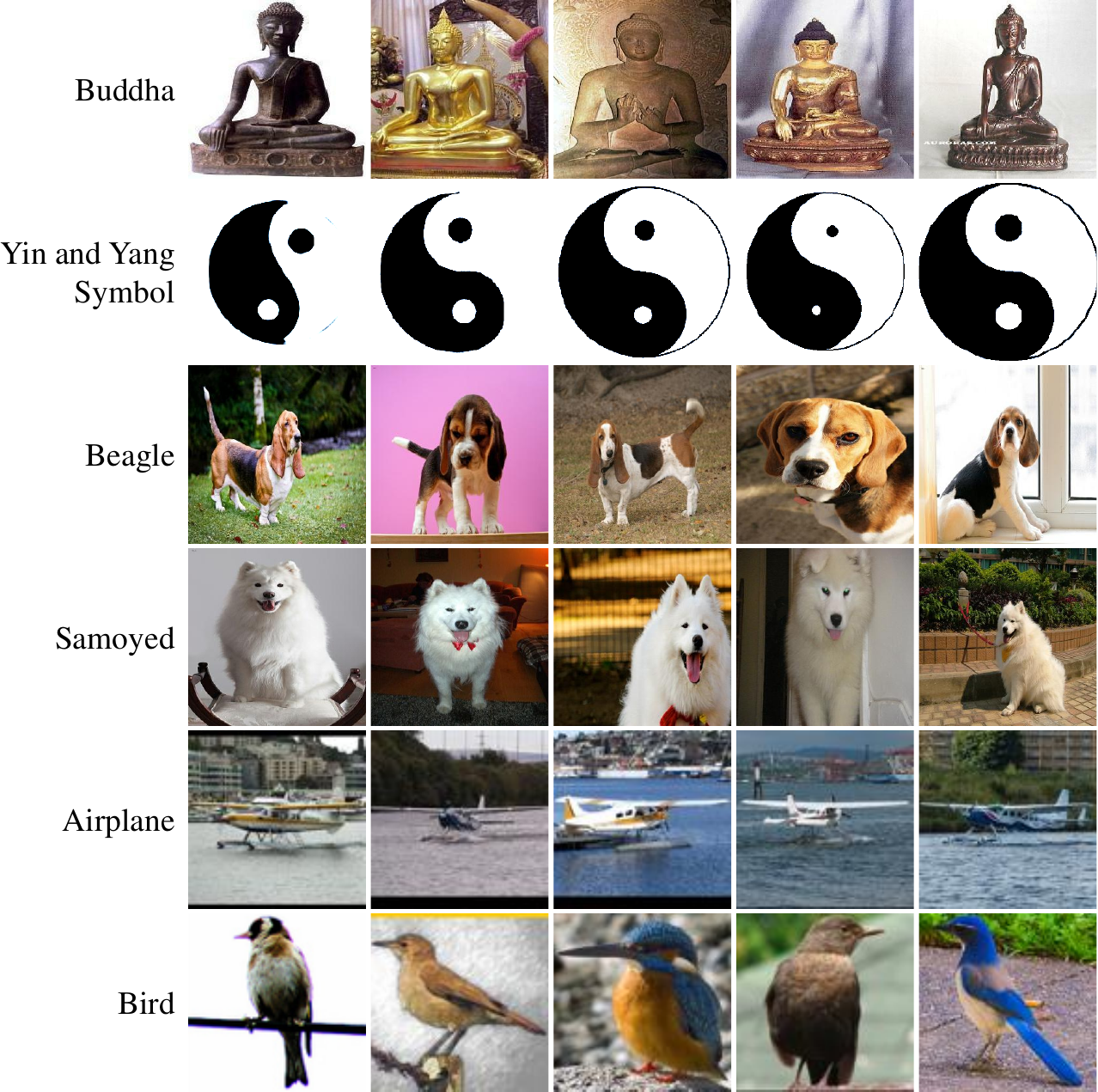}
    \caption{\textbf{Visualization of zero-shot classification results}. We provide the top-5 predictions of our proposed CLIP-PGS$_{0.3}$. The first two rows report examples from Caltech101~\cite{caltech101}, the next two rows highlight samples from OxfordPets~\cite{pets}, and the final two rows present results from STL10~\cite{stl10}.}
    \label{fig:appendix_cls}
\end{figure}

\begin{figure}[t]
    \centering
    \includegraphics[width=1.0\linewidth]{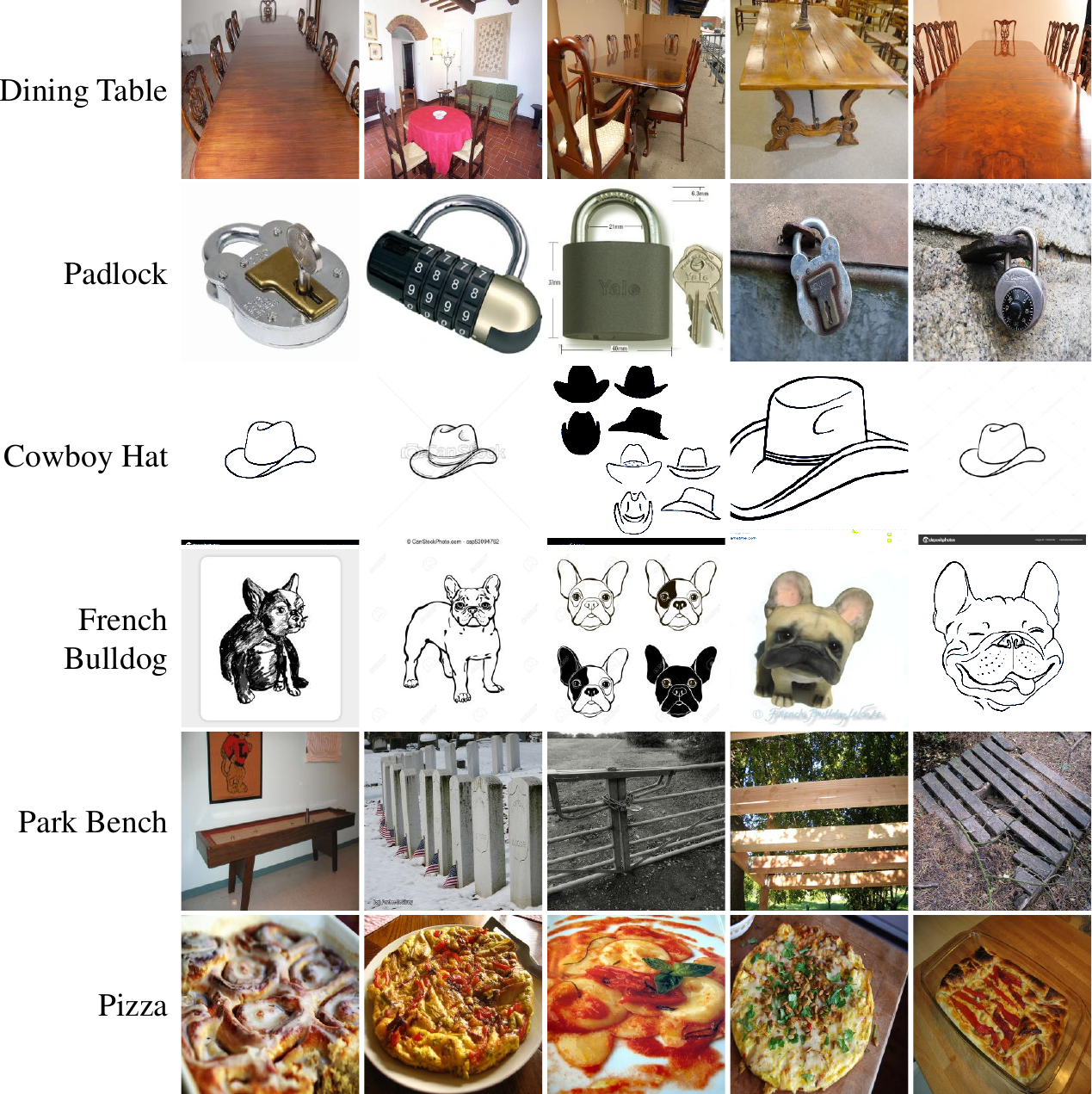}
    \caption{\textbf{Visualization of zero-shot robustness results}. We provide the top-5 predictions of our proposed CLIP-PGS$_{0.3}$. The first two rows report examples from ImageNet-1K~\cite{deng2009imagenet}, the next two rows highlight samples from ImageNet-R~\cite{imagenet-r}, and the final two rows present results from ImageNet-O~\cite{imagenet-ao}.}
    \label{fig:appendix_rob}
\end{figure}

\noindent\textbf{Classification Results.}
This section highlights the qualitative performance of CLIP-PGS in zero-shot classification and robustness tasks, showcasing its adaptability and generalization.
For zero-shot classification (\cref{fig:appendix_cls}), examples from diverse datasets demonstrate the model's accurate alignment of visual and textual representations, adapting effectively to varied categories and contexts.
For zero-shot robustness (\cref{fig:appendix_rob}), results from robustness-focused datasets illustrate CLIP-PGS's resilience in handling distributional shifts while maintaining high prediction quality.

\begin{figure}[t]
    \centering
    \includegraphics[width=1.0\linewidth]{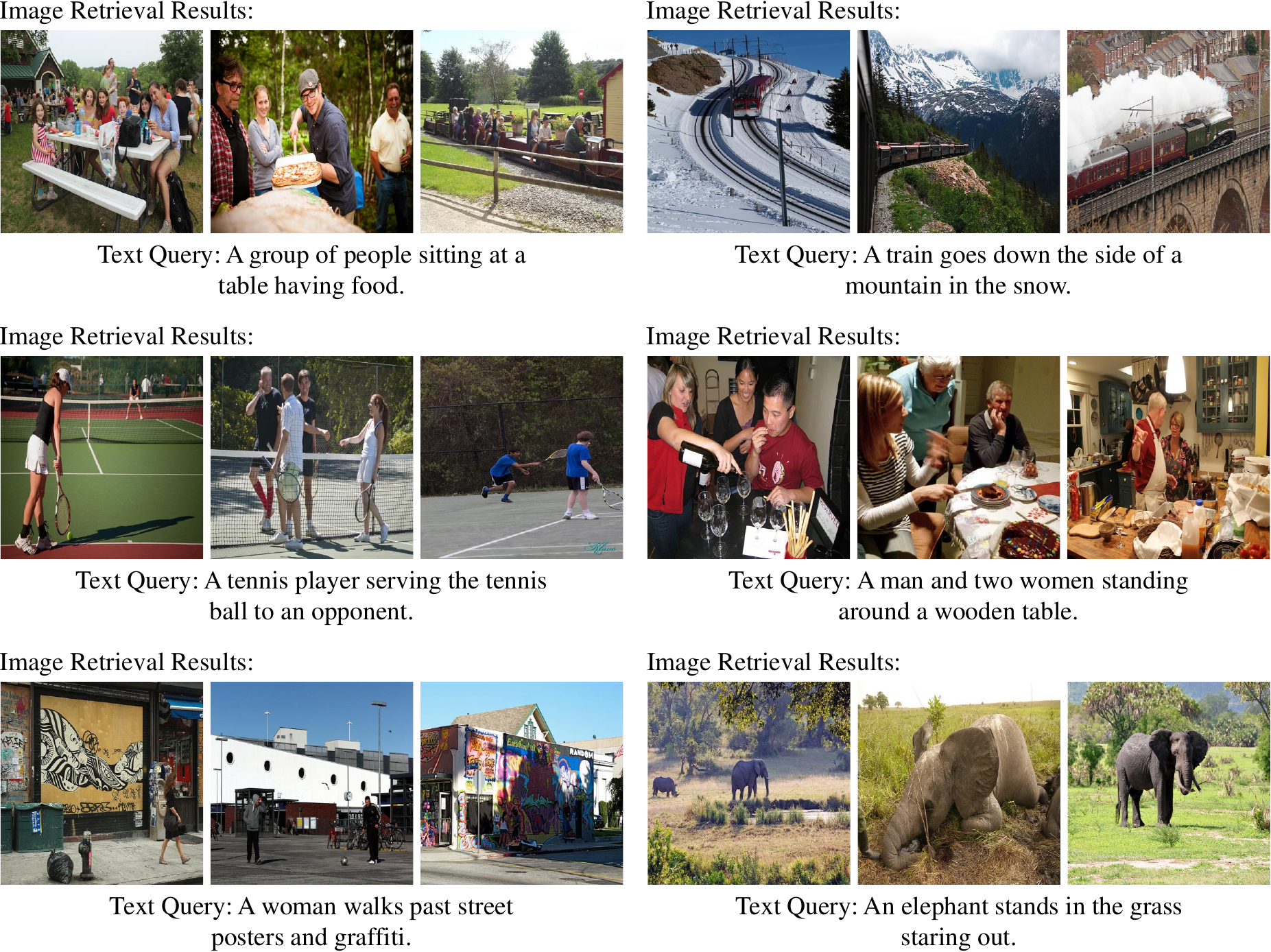}
    \caption{\textbf{Visualization of zero-shot text retrieval}. We provide the top-3 predictions of our proposed CLIP-PGS$_{0.3}$. Examples are from the retrieval dataset MS-COCO~\cite{ms-coco}.}
    \label{fig:appendix_ret_t2i}
\end{figure}

\begin{figure}[t]
    \centering
    \includegraphics[width=1.0\linewidth]{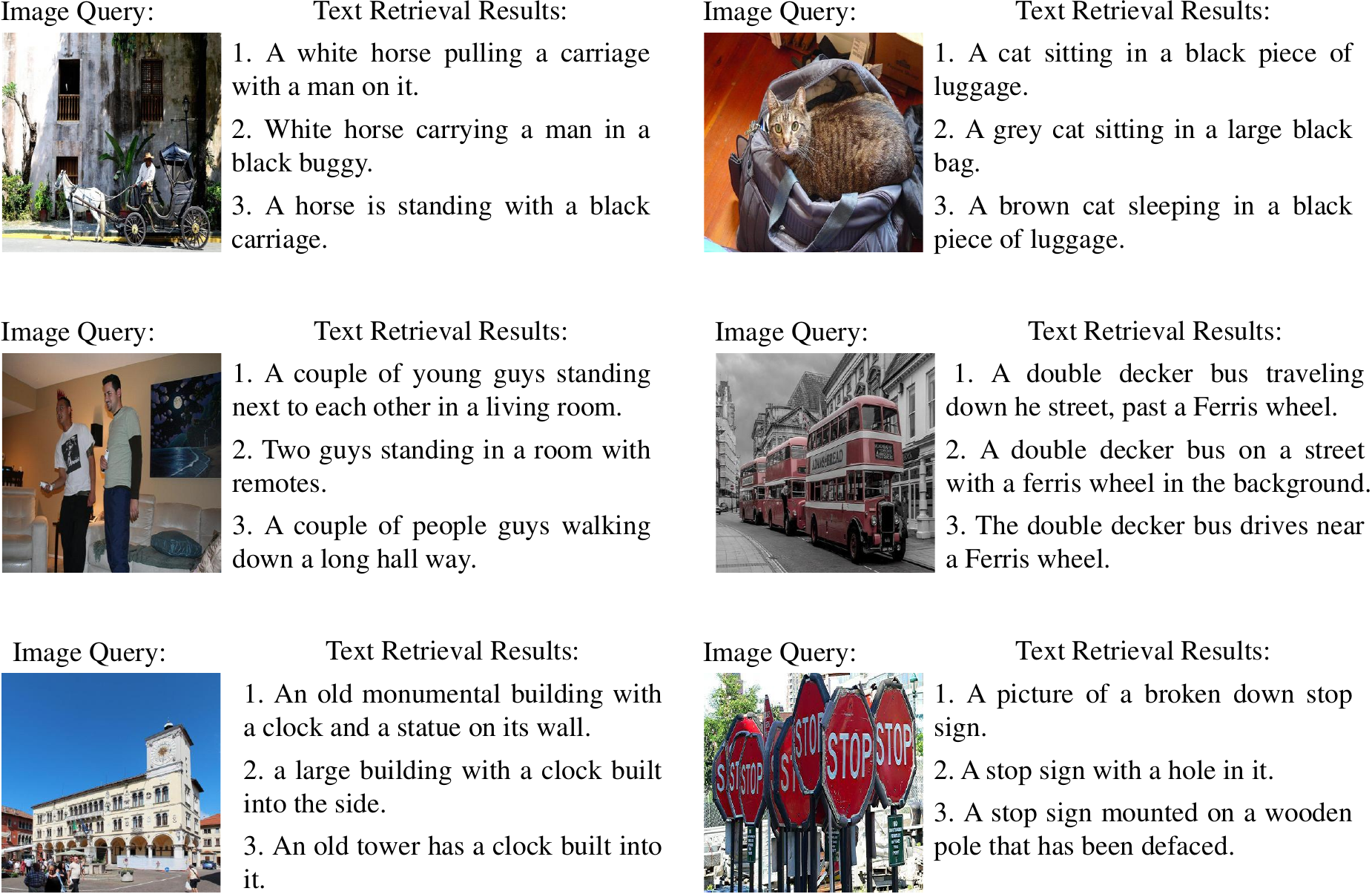}
    \caption{\textbf{Visualization of zero-shot image retrieval}. We provide the top-3 predictions of our proposed CLIP-PGS$_{0.3}$. Examples are from the retrieval dataset MS-COCO~\cite{ms-coco}.}
    \label{fig:appendix_ret_i2t}
\end{figure}

\noindent\textbf{Retrieval Results.}
This section evaluates CLIP-PGS on zero-shot text and image retrieval tasks using the MS-COCO~\cite{ms-coco} dataset.
For text retrieval (\cref{fig:appendix_ret_t2i}), the model retrieves highly relevant images for given text queries, showcasing precise alignment between textual and visual content.
For image retrieval (\cref{fig:appendix_ret_i2t}), CLIP-PGS effectively links input images to their corresponding textual descriptions, demonstrating robust cross-modal associations even in complex scenarios.

\clearpage


\end{document}